\documentclass[manuscript]{acmart}

\usepackage{xcolor}
\AtBeginDocument{%
  }

\usepackage{booktabs}       
\usepackage{graphicx}
\usepackage{multirow}
\usepackage{tabularx} 
\usepackage{amsmath}
\usepackage{subcaption}
\usepackage{rotating}
\usepackage{array}
\usepackage{wrapfig}
\usepackage{amsthm}
\usepackage{enumitem}
\newtheorem{definition}{Definition}

\usepackage{tikz}
\usetikzlibrary{shapes.multipart, positioning, calc, arrows.meta}

\usepackage[edges]{forest}

\definecolor{c1}{RGB}{127,127,127} 
\definecolor{c2}{RGB}{68,114,196} 
\definecolor{c3}{RGB}{218,227,243} 

\tikzset{
    tree leaf connection/.style={
        draw=c2,
        line width=0.7pt,
        dash pattern=on 2.2pt off 1.4pt,
        -{Latex[length=1.6mm,width=1.0mm]},
        shorten <=0pt,
        shorten >=0pt,
    },
    tree leaf bidirectional connection/.style={
        draw=c2,
        line width=0.7pt,
        {Latex[length=1.6mm,width=1.0mm]}-{Latex[length=1.6mm,width=1.0mm]},
        shorten <=0pt,
        shorten >=0pt,
    },
    tree leaf connection dot/.style={
        circle,
        draw=c2,
        fill=white,
        line width=0.7pt,
        inner sep=0pt,
        minimum size=3pt,
    },
    tree leaf bidirectional connection dot/.style={
        tree leaf connection dot,
        fill=c2,
    },
}
\newcommand{\DrawTreeLeafArrow}[3][]{%
    \node[tree leaf connection dot] (#2Dot) at (#2) {};%
    \node[tree leaf connection dot] (#3Dot) at (#3) {};%
    \draw[tree leaf connection,#1] (#2Dot.east) -- (#3Dot.west);%
}
\newcommand{\DrawTreeLeafBidirectionalArrow}[3][]{%
    \node[tree leaf bidirectional connection dot] (#2Dot) at (#2) {};%
    \node[tree leaf bidirectional connection dot] (#3Dot) at (#3) {};%
    \draw[tree leaf bidirectional connection,#1] (#2Dot.east) -- (#3Dot.west);%
}

\definecolor{bluegray}{HTML}{5B6F84}

\newcommand{\TreeDefinition}[1]{\textcolor{c1}{Definition~\ref{#1}}}
\newcommand{\TreeCitation}[1]{\begingroup\hypersetup{citecolor=c1}\textcolor{c1}{\citep{#1}}\endgroup}

\usepackage{soul}
\sethlcolor{yellow}
\soulregister\citep7
\soulregister\ref7

\usepackage{tcolorbox}
\tcbuselibrary{breakable}

\makeatletter
\newcounter{subsubsubsection}[subsubsection]
\renewcommand{\thesubsubsubsection}{\thesubsubsection.\alph{subsubsubsection}}

\newcommand{\subsubsubsection}[1]{%
  \vspace*{0.5\baselineskip}
  \refstepcounter{subsubsubsection}%
  \noindent
  {\ACM@NRadjust{\@subsubsecfont \thesubsubsubsection\quad #1.}}%
}
\makeatother

\begin{document}

\title{Bridging Distribution Shift and AI Safety: Conceptual and Methodological Synergies
}

\author{Chenruo Liu}
\authornote{Both authors contributed equally to this research.}
\email{cl7758@nyu.edu}
\affiliation{%
  \institution{Center for Data Science, New York University}
  \city{New York}
  \state{New York}
  \country{USA}
}

\author{Kenan Tang}
\authornotemark[1]
\email{kenantang@ucsb.edu}
\affiliation{%
  \institution{Computer Science Department, University of California, Santa Barbara}
  \city{Santa Barbara}
  \state{California}
  \country{USA}
}

\author{Yao Qin}
\email{yaoqin@ucsb.edu}
\affiliation{%
  \institution{Department of Electrical and Computer Engineering, University of California, Santa Barbara}
  \city{Santa Barbara}
  \state{California}
  \country{USA}
}

\author{Qi Lei}
\email{ql518@nyu.edu}
\affiliation{%
  \institution{Courant Institute for Mathematical Sciences \& Center for Data Science, New York University}
  \city{New York}
  \state{New York}
  \country{USA}
}

\renewcommand{\shortauthors}{Liu et al.}

\begin{abstract}
This paper bridges distribution shift and AI safety through a comprehensive analysis of their conceptual and methodological synergies. While prior discussions often focus on narrow cases or informal analogies, we establish two types of connections between specific causes of distribution shift and fine-grained AI safety issues: (1) methods addressing a specific shift type can help achieve corresponding safety goals, or (2) certain shifts and safety issues can be formally reduced to each other, enabling mutual adaptation of their methods. Our findings provide a unified perspective that encourages deeper integration between distribution shift and AI safety research.
\end{abstract}



\settopmatter{printacmref=false}

\maketitle

\section{Introduction}
\label{sec:introduction}

Machine learning (ML) models have achieved remarkable success across complex tasks in various fields, including high-stakes applications such as autonomous driving, healthcare, finance, and legal domains~\cite{grigorescu2020survey,shailaja2018machine,dixon2020machine,surden2021machine}. Traditionally, ML models are trained on historically collected data and subsequently deployed in a test domain. While these ML models succeed in the case where the training and testing data share the same distribution, performance drops are expected when distribution shifts from training to test domains~\citep{koh2021wilds, cai2025diagnosing}. 

Beyond performance concerns, the failure of ML models to robustly handle distribution shifts in real-world settings can lead to severe safety issues, such as malfunctions in autonomous driving systems~\citep{stoler2024safeshift} and errors in clinical systems~\citep{ma2022test}. In more recent examples, despite substantial efforts in alignment research, large language models (LLMs) and large multimodal models still suffer from outputting inappropriate content under distribution shift~\citep{wang2023robustness, jeong2025playing}. As artificial intelligence (AI) systems continue to evolve, minimizing their safety risks has emerged as a critical priority~\citep{amodei2016concrete}. Consequently, distribution shift, which is inherently linked to AI safety, has become a central research challenge.

A deep understanding of the relationship between distribution shift and AI safety is undoubtedly beneficial for the safe deployment of ML models in real-world applications. However, although previous studies have frequently mentioned the connection between these two fields~\citep{amodei2016concrete, li2023trustworthy, ji2023ai, lu2025alignment, fang2025trustworthy}, they exhibit several limitations. First, they focus narrowly on the relationship between distribution shift and a single aspect of AI safety (e.g., robustness~\citep{amodei2016concrete}, trustworthiness~\citep{fang2025trustworthy, li2023trustworthy}, or alignment~\citep{lu2025alignment, ji2023ai}) but lack a more comprehensive framework. Second, these studies often treat both distribution shift and the targeted AI safety aspect as general concepts, rather than breaking them down into specific categories~\citep{amodei2016concrete}. Third, existing work discusses the connection between distribution shift and AI safety as an informal analogy without identifying the mathematical relationships among their rigorous definitions. As a result, there is still a lack of detailed exploration into the connections between specific topics in distribution shift and AI safety.

\begin{figure*}[!pt]
    \centering
    \begin{tikzpicture}
        \node[inner sep=0pt, anchor=north west] (distributionTree) at (0,0) {%
        \resizebox{0.41\textwidth}{!}{
        \begin{forest}
            forked edges,
            for tree={
                grow=east,
                reversed=true,
                anchor=base west,
                parent anchor=east,
                child anchor=west,
                base=center,
                font=\large,
                rectangle,
                draw=gray,
                rounded corners,
                align=center,
                text centered,
                minimum width=4em,
                edge+={c2, line width=1pt},
                s sep=3pt,
                inner xsep=2pt,
                inner ysep=3pt,
                line width=1pt,
                ver/.style={rotate=90, child anchor=north, parent anchor=south, anchor=center},
            },
            where level=2{text width=5em,font=\normalsize,}{},
            where level=3{text width=10.5em,font=\normalsize,}{},
            where level=4{text width=22em,font=\normalsize,}{},
            [
                \textbf{Causes of Distribution Shift}, ver, line width=0.7mm, text=white, fill=c1, draw=c1,
                [
                    $P_X$ \textbf{changes}, ver, fill=c2, draw=c2, text=white, line width=1pt,
                        [
                        Selection\\Bias, fill=c3, draw=c2, line width=1pt,
                        [
                            Individual Selection Bias\\
                            \TreeDefinition{def:individual selection bias} \textcolor{c1}{(\S~\ref{sec:individual-selection-bias})}\\
                            \TreeCitation{recht2019imagenet, sahiner2023data, geirhos2018imagenettrained, nam2020learning, qin2023effective}
                            , draw=c2, line width=1pt, name=distIndividualSelection,
                        ]
                        [
                            Group Selection Bias\\
                            \TreeDefinition{def:group selection bias} \textcolor{c1}{(\S~\ref{sec:group-selection-bias})}\\
                            \TreeCitation{karkkainen2021fairface, ni-etal-2019-justifying, Sagawa2020Distributionally, idrissi2022simple, krueger2021out}
                            , draw=c2, line width=1pt, name=distGroupSelection,
                        ]
                    ]
                    [
                        Spurious\\Correlation, fill=c3, draw=c2, line width=1pt
                        [
                            Environmental Change\\
                            \TreeDefinition{def:Environmental Change} \textcolor{c1}{(\S~\ref{sec:environmental-change})}\\
                            \TreeCitation{peters2016causal, arjovsky2019invariant, wald2021calibration, pan2010domain, zellinger2017central}
                            , draw=c2, line width=1pt, name=distEnvironmental,
                        ]
                        [
                            Content Change\\
                            \TreeDefinition{def:Content Change} \textcolor{c1}{(\S~\ref{subsec:Content change})}\\
                            \TreeCitation{zhang2021deep, zhang2022towards, teney2022evading, chen2023project, fumero2023leveraging}
                            , draw=c2, line width=1pt, name=distContent,
                        ]
                    ]
                ]
                [
                    $P_Y$ \textbf{changes}, ver, fill=c2, draw=c2, text=white, line width=1pt
                    [
                        Label\\Shift, fill=c3, draw=c2, line width=1pt,
                        edge path={
                            \noexpand\path[\forestoption{edge}]
                            (!u.south |- .west) -- (.west)
                            \forestoption{edge label};
                        },
                        [
                            Generalized Label Shift\\
                            \TreeDefinition{def:Generalized Label Shift} \textcolor{c1}{(\S~\ref{sec:generalized-label-shift})}\\
                            \TreeCitation{tachet2020domain, gong2016domain, shui2021aggregating, mahajan2021domain, wu2023prominent}
                            , draw=c2, line width=1pt, name=distGeneralizedLabel,
                        ]
                        [
                            Open-Set Label Shift\\
                            \TreeDefinition{def:OSLS} \textcolor{c1}{(\S~\ref{sec:open-set-label-shift})}\\
                            \TreeCitation{garg2022domain, ge2017generative, you2019universal, fu2020learning, saito2020universal}
                            , draw=c2, line width=1pt, name=distOpenSetLabel,
                        ]
                    ]
                ]
            ]
        \end{forest}
        }%
        };
        \node[inner sep=0pt, anchor=north west] (safetyTree) at (7.2cm,0.46cm) {%
        \resizebox{0.52\textwidth}{!}{
        \begin{forest}
            forked edges,
            for tree={
                grow=west,
                reversed=false,
                anchor=base east,
                parent anchor=west,
                child anchor=east,
                base=center,
                font=\large,
                rectangle,
                draw=gray,
                rounded corners,
                align=center,
                text centered,
                minimum width=4em,
                edge+={c2, line width=1pt},
                s sep=3pt,
                inner xsep=2pt,
                inner ysep=3pt,
                line width=1pt,
                ver/.style={rotate=270, child anchor=north, parent anchor=south, anchor=center},
            },
            where level=2{text width=6.5em,font=\normalsize,}{},
            where level=3{text width=16em,font=\normalsize,}{},
            where level=4{text width=16em,font=\normalsize,}{},
            [
                \textbf{AI Safety}, ver, line width=0.7mm, fill=c1, draw=c1, text=white
                [
                    \textbf{Security}, ver, fill=c2, draw=c2, text=white, line width=1pt,
                    [
                        Security\\Attack, fill=c3, draw=c2, line width=1pt,
                        edge path={
                            \noexpand\path[\forestoption{edge}]
                            (!u.south |- .east) -- (.east)
                            \forestoption{edge label};
                        },
                        [
                            Backdoor Poisoning Attack\\
                            \TreeDefinition{def:Backdoor Poisoning Attack} \textcolor{c1}{(\S~\ref{sec:environmental-change-security})}\\
                            \TreeCitation{gu2017badnets, gao2020backdoor, li2023backdoor, chen2017targeted, turner2019label}
                            , draw=c2, line width=1pt, name=safetyBackdoor,
                        ]
                    ]
                ]
                [
                    \textbf{Fairness}, ver, fill=c2, draw=c2, text=white, line width=1pt,
                    [
                        Group\\Fairness, fill=c3, draw=c2, line width=1pt,
                        edge path={
                            \noexpand\path[\forestoption{edge}]
                            (!u.south |- .east) -- (.east)
                            \forestoption{edge label};
                        },
                        [
                            Risk Parity\\
                            \TreeDefinition{def:risk parity} \textcolor{c1}{(\S~\ref{sec:group-selection-bias-fairness})}
                            , draw=c2, line width=1pt, name=safetyRiskParity,
                        ]
                        [
                            Test Fairness\\
                            \TreeDefinition{def:TestFairness} \textcolor{c1}{(\S~\ref{sec:environmental-change-fairness})}
                            , draw=c2, line width=1pt, name=safetyTestFairness,
                        ]
                        [
                            Well-Calibration\\
                            \TreeDefinition{def:WellCalibration} \textcolor{c1}{(\S~\ref{sec:environmental-change-fairness})}
                            , draw=c2, line width=1pt, name=safetyWellCalibration,
                        ]
                        [
                            Demographic Parity\\
                            \TreeDefinition{def:DemographicParity} \textcolor{c1}{(\S~\ref{sec:environmental-change-fairness})}
                            , draw=c2, line width=1pt, name=safetyDemographicParity,
                        ]
                        [
                            Equalized Odds\\
                            \TreeDefinition{def:EqualizedOdds} \textcolor{c1}{(\S~\ref{sec:environmental-change-fairness2})}
                            , draw=c2, line width=1pt, name=safetyEqualizedOdds,
                        ]
                    ]
                ]
                [
                    \textbf{Trustworthiness}, ver, fill=c2, draw=c2, text=white, line width=1pt,
                    [
                        Interpretability, fill=c3, draw=c2, line width=1pt,
                        [
                            Disentangled Representation Learning\\
                            \TreeDefinition{def:Disentangled Representation} \textcolor{c1}{(\S~\ref{sec:environmental-change-interpretability} and \S~\ref{sec:content-change-trustworthiness})}
                            , draw=c2, line width=1pt, name=safetyDisentangled,
                        ]
                    ]
                    [
                        Alignment, fill=c3, draw=c2, line width=1pt,
                        [
                            Inner Alignment\\
                            \TreeDefinition{def:Inner and Outer Alignment}  \textcolor{c1}{(\S~\ref{sec:environmental-change-alignment})}
                            , draw=c2, line width=1pt, name=safetyInnerAlignment,
                        ]
                    ]
                    [
                        Uncertainty\\Quantification, fill=c3, draw=c2, line width=1pt,
                        [
                            Uncertainty Quantification\\
                            \TreeDefinition{def:Uncertainty Quantification}  \textcolor{c1}{(\S~\ref{sec:uncertainty-quantification})}
                            , draw=c2, line width=1pt, name=safetyUncertainty,
                        ]
                    ]
                ]
                [
                    \textbf{Democracy}, ver, fill=c2, draw=c2, text=white, line width=1pt,
                    [
                        Data\\Pruning, fill=c3, draw=c2, line width=1pt,
                        edge path={
                            \noexpand\path[\forestoption{edge}]
                            (!u.south |- .east) -- (.east)
                            \forestoption{edge label};
                        },
                        [
                            Data Pruning\\
                            \TreeDefinition{def:Data Pruning} \textcolor{c1}{(\S~\ref{sec:democracy})}\\
                            \TreeCitation{maharana2024mathbbd, ben2024distilling, caldeira2025diffprob, meding2022trivial, Coleman2020Selection}
                            , draw=c2, line width=1pt, name=safetyDataPruning,
                        ]
                    ]
                ]
            ]
        \end{forest}
        }%
        };
        \coordinate (distIndividualSelectionArrowStart) at ($(distributionTree.north east)+(-0.033cm,-0.588cm)$);
        \coordinate (distGroupSelectionArrowStart) at ($(distributionTree.north east)+(-0.033cm,-1.835cm)$);
        \coordinate (distEnvironmentalBackdoorArrowStart) at ($(distributionTree.north east)+(-0.033cm,-2.701cm)$);
        \coordinate (distEnvironmentalTestArrowStart) at ($(distributionTree.north east)+(-0.033cm,-2.854cm)$);
        \coordinate (distEnvironmentalCalibrationArrowStart) at ($(distributionTree.north east)+(-0.033cm,-3.006cm)$);
        \coordinate (distEnvironmentalDemographicArrowStart) at ($(distributionTree.north east)+(-0.033cm,-3.158cm)$);
        \coordinate (distEnvironmentalDisentangledArrowStart) at ($(distributionTree.north east)+(-0.033cm,-3.310cm)$);
        \coordinate (distEnvironmentalAlignmentArrowStart) at ($(distributionTree.north east)+(-0.033cm,-3.463cm)$);
        \coordinate (distContentArrowStart) at ($(distributionTree.north east)+(-0.033cm,-4.329cm)$);
        \coordinate (distGeneralizedLabelArrowStart) at ($(distributionTree.north east)+(-0.033cm,-5.576cm)$);
        \coordinate (distOpenSetLabelArrowStart) at ($(distributionTree.north east)+(-0.033cm,-6.823cm)$);
        \coordinate (safetyBackdoorArrowEnd) at ($(safetyTree.north west)+(-0.030cm,-0.579cm)$);
        \coordinate (safetyRiskParityArrowEnd) at ($(safetyTree.north west)+(-0.030cm,-1.635cm)$);
        \coordinate (safetyTestFairnessArrowEnd) at ($(safetyTree.north west)+(-0.030cm,-2.547cm)$);
        \coordinate (safetyWellCalibrationArrowEnd) at ($(safetyTree.north west)+(-0.030cm,-3.460cm)$);
        \coordinate (safetyDemographicParityArrowEnd) at ($(safetyTree.north west)+(-0.030cm,-4.360cm)$);
        \coordinate (safetyEqualizedOddsFromGeneralizedLabelArrowEnd) at ($(safetyTree.north west)+(-0.030cm,-5.270cm)$);
        \coordinate (safetyDisentangledFromEnvironmentalArrowEnd) at ($(safetyTree.north west)+(-0.030cm,-6.060cm)$);
        \coordinate (safetyDisentangledFromContentArrowEnd) at ($(safetyTree.north west)+(-0.030cm,-6.299cm)$);
        \coordinate (safetyInnerAlignmentArrowEnd) at ($(safetyTree.north west)+(-0.030cm,-7.087cm)$);
        \coordinate (safetyUncertaintyArrowEnd) at ($(safetyTree.north west)+(-0.030cm,-7.995cm)$);
        \coordinate (safetyDataPruningArrowEnd) at ($(safetyTree.north west)+(-0.030cm,-9.095cm)$);
        \DrawTreeLeafBidirectionalArrow{distIndividualSelectionArrowStart}{safetyDataPruningArrowEnd}
        \DrawTreeLeafArrow{distGroupSelectionArrowStart}{safetyRiskParityArrowEnd}
        \DrawTreeLeafBidirectionalArrow{distEnvironmentalBackdoorArrowStart}{safetyBackdoorArrowEnd}
        \DrawTreeLeafArrow{distEnvironmentalTestArrowStart}{safetyTestFairnessArrowEnd}
        \DrawTreeLeafArrow{distEnvironmentalCalibrationArrowStart}{safetyWellCalibrationArrowEnd}
        \DrawTreeLeafArrow{distEnvironmentalDemographicArrowStart}{safetyDemographicParityArrowEnd}
        \DrawTreeLeafArrow{distEnvironmentalDisentangledArrowStart}{safetyDisentangledFromEnvironmentalArrowEnd}
        \DrawTreeLeafArrow{distEnvironmentalAlignmentArrowStart}{safetyInnerAlignmentArrowEnd}
        \DrawTreeLeafArrow{distContentArrowStart}{safetyDisentangledFromContentArrowEnd}
        \DrawTreeLeafArrow{distGeneralizedLabelArrowStart}{safetyEqualizedOddsFromGeneralizedLabelArrowEnd}
        \DrawTreeLeafArrow{distOpenSetLabelArrowStart}{safetyUncertaintyArrowEnd}
    \end{tikzpicture}
    \caption{Connections between distribution shift causes and AI safety issues. In each leaf node, the cited definition number indicates where the concept is formally defined, and the section number in parentheses indicates the section where the connection is discussed in detail. One-directional arrows: Methods addressing a specific cause of distribution shift can also achieve a particular safety goal. Bidirectional solid arrows: Due to inherent consistency between a certain distribution shift cause and a corresponding safety issue, the two can be reduced from one another and methods derived for them can be mutually adapted. For one-directional connections, we list five representative references under the corresponding distribution shift cause. For bidirectional connections, we list five representative references under both the distribution shift cause and the associated safety issue. Although environmental change is linked to multiple safety issues, we show only five representative references here. More can be found in the corresponding sections.}
    \label{fig:Connection All}
    \vspace{-0.6em}
\end{figure*}

In this paper, we offer the first comprehensive, one-to-one aligned, and mathematically grounded analysis of the relationship between distribution shift and AI safety (Figure \ref{fig:Connection All}). On the one hand, according to the underlying causes, we examine distribution shifts from three main categories: \textit{selection bias}~\citep{shimodaira2000improving, santurkar2020breeds}, \textit{spurious correlation}~\citep{simon1954spurious, ye2024spurious}, and \textit{label shift}~\citep{lipton2018detecting, tachet2020domain, garg2022domain} and six more fine-grained categories. On the other hand, we focus on the following categories of AI safety issues: \textit{AI security}~\citep{sarker2021ai}, \textit{AI fairness}~\citep{zhou2020towards}, \textit{AI trustworthiness}~\citep{kaur2022trustworthy}, and \textit{AI democracy}~\citep{seger2023democratising}. For each specific topic discussed within these main categories, we provide its rigorous definition. Next, we systematically identify connections between specific distribution shift causes and concrete AI safety issues by analyzing the underlying mathematical relationships between their definitions. Based on the identified connections, we then review the relevant literature to establish methodological links between these specific topics.

Our analysis exhibits three characteristics that distinguish it from previous work:
\begin{itemize}[itemsep=0.5pt, parsep=0.5pt, topsep=0.5pt]
    \item \textbf{Comprehensive}: We cover a diverse range of aspects of distribution shifts and AI safety, providing a more comprehensive discussion of the relationship between these two fields than previous specialized surveys.
    \item \textbf{One-to-one aligned}: We directly align each specific distribution shift cause with relevant concrete AI safety issues to enable a clear, one-to-one comparison. The comparison leads to actionable future directions.
    \item \textbf{Mathematically grounded}: The connections we identify are rooted in rigorous mathematical relationships, as shown in the ``Definition-Level Connection'' boxes, which guide our subsequent review of the literature.
\end{itemize}

Building on these features, our work reveals two key relationships between distribution shifts and AI safety. First, methods addressing a specific cause of distribution shift can achieve a concrete AI safety goal (indicated by one-directional arrows in Figure \ref{fig:Connection All}). Second, methods designed for a specific distribution shift cause and a concrete AI safety issue can be mutually adapted because of their definitional consistency (indicated by bidirectional solid arrows in Figure \ref{fig:Connection All}). These two types of connections highlight shared research interests and create opportunities for methodological exchange across distribution shift and AI safety. Our findings establish a more unified perspective on distribution shifts and AI safety, fostering closer collaboration between these two research communities. 

The remainder of this paper is organized as follows. Section \ref{sec:background} introduces notations used throughout the paper and provides a topic overview. In Sections \ref{sec:selection-bias}, \ref{sec:spurious-correlation}, and \ref{sec:label-shift}, we examine how the specific causes of distribution shift within the main categories are connected to various concrete AI safety issues. Finally, Section \ref{sec:conclusions} concludes the paper.

\section{Background}
\label{sec:background}

This section introduces the notations, categorizations, and logical structure that underpin our systematic analysis of the connections between distribution shift and AI safety. We first present the necessary notations with a brief introduction to distribution shift, and then offer an overview of the main categories under both distribution shift and AI safety.

\subsection{Notations}
\label{sec:notations}

Since this paper is organized from the perspective of distribution shift, we begin by introducing common notations for distribution shift. Unless otherwise specified, we consider the task of predicting the label \(Y\) from the input variable \(X\). The prediction model is denoted by \(f\). When necessary, we view \(f\) as a composition of a feature extractor \(\Phi\) and a prediction head (also called a classifier) \(\omega\), i.e., \(f = \omega \circ \Phi\).

Within the context of distribution shift, we introduce a domain (also known as environment\footnote{In this paper, domain and environment are used interchangeably. We will use the term that fits the specific context or convention.}) variable \(E\) indexing any possible domain. It is typically assumed that the training data are drawn from a source domain \(e_s\) with joint distribution \(P(X,Y\mid E=e_s)\), while the test data are drawn from a target domain \(e_t\) with joint distribution \(P(X,Y\mid E=e_t)\). Under this formulation, the distribution shift between the training and test domains can be expressed as
\[
P(X,Y\mid E=e_s) \neq P(X,Y\mid E=e_t).
\]
Due to this distribution shift, the model learned by minimizing the population risk in the training domain \(\mathcal{R}_{e_s}(f) = \mathbb{E}\left[\,l\bigl(f(X),Y\bigr)\mid E=e_s\right]\), may suffer significant performance degradation when evaluated on the test data, because it is not optimized for the test domain's risk \(\mathcal{R}_{e_t}(f)\). Since the true population distribution is unavailable in practice, we often learn the model \(f\) via empirical risk minimization (ERM), i.e., we approximate \(\mathcal{R}_{e_s}(f)\) with the empirical risk \(\hat{\mathcal{R}}_{e_s}(f) = \frac{1}{n}\sum_{i=1}^{n} l\bigl(f(x_{e_s}^{(i)}),y_{e_s}^{(i)}\bigr)\), where each \((x_{e_s}^{(i)}, y_{e_s}^{(i)})\) is independently and identically sampled from \(P(X,Y\mid E=e_s)\).

Sometimes, there are multiple training domains with different joint distributions. In this case, distribution shift means that the test domain's distribution differs from that of any training domain. Moreover, if a single environment or domain is considered as a mixture of different groups, we use the variable \(G\) to denote the group label. 

\subsection{Topic Overview}
\label{sec:topic-overview}

\begin{figure}[htbp]
  \centering
  \includegraphics[width=0.8\textwidth]{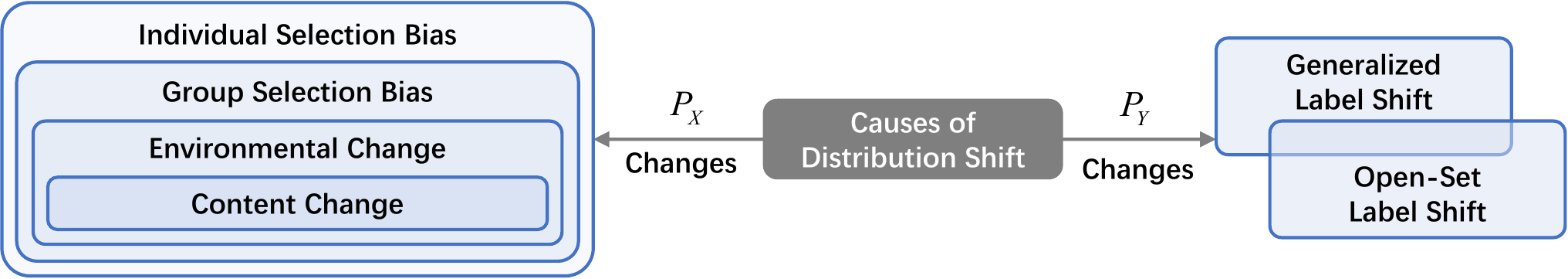} 
  \caption{Illustration of the inclusion relationships among different causes of distribution shift.}

  \label{fig:Relationship Distribution Shift}
\end{figure}

Distribution shift can arise from a wide range of factors. In this paper, according to the underlying causes that lead to distribution shift, we classify it into three main categories: selection bias, spurious correlation, and label shift (Figure \ref{fig:Connection All}). In general, selection bias refers to the bias induced by the selection process of input \(X\) in the training domain~\citep{shimodaira2000improving, santurkar2020breeds}. Spurious correlation refers to the non-causal correlation present in the training data that changes in the test data~\citep{simon1954spurious, ye2024spurious}. Label shift focuses on the shift in the distribution of label \(Y\) between the training and test domains~\citep{lipton2018detecting, tachet2020domain, garg2022domain}. Within each main category, we further consider the specific distribution shift causes. Figure \ref{fig:Relationship Distribution Shift} illustrates the inclusion relationships among these causes, according to the definitions in respective sections (Figure \ref{fig:Connection All}).

For each cause of distribution shift, we examine relevant AI safety issues  based on the mathematical relationships among the definitions. The concrete AI safety issues under consideration cover four main categories: security, fairness, trustworthiness, and democracy (Figure \ref{fig:Connection All}). AI security refers to using AI to protect systems and data from cyber-threats~\citep{sarker2021ai,uddin2025generative}. AI fairness ensures that models do not exhibit prejudice against individuals or groups~\citep{zhou2020towards, he2025fairness}. AI trustworthiness requires that AI systems meet user expectations in a verifiable manner~\citep{kaur2022trustworthy, fang2025trustworthy}. AI democracy involves the equitable spread of AI applications to enable more people to access and benefit from the technology~\citep{mushkani2025position, seger2023democratising}. While there is a broad spectrum of other AI safety issues, these four are chosen as the primary focus of our survey.

\section{Selection Bias}
\label{sec:selection-bias}

Selection bias occurs when the process of selecting training data leads to a dataset that does not accurately represent the target population. This misrepresentation creates a discrepancy between the training and test distributions, often resulting in a drop in performance during deployment \cite{shi2023effective, liu2025detecting, liu2024fast}. For instance, consider an image classifier designed to recognize everyday objects like chairs and bikes. If the model is trained on real-world photographs but later evaluated on paintings of these objects, it may fail to generalize effectively. While humans can easily recognize objects despite stylistic changes, the classifier struggles with the distribution shift, leading to lower accuracy on the test set.

In the following subsections, we further categorize selection bias into individual selection bias (Section \ref{sec:individual-selection-bias}) and group selection bias (Section \ref{sec:group-selection-bias}). Then, we connect them to AI safety issues based on mathematical definitions.

\subsection{Individual Selection Bias}
\label{sec:individual-selection-bias}

\begin{figure}[htbp]
  \centering
  \hypertarget{selection-bias}{}
  \begin{subfigure}[t]{0.35\textwidth}
    \centering
    \includegraphics[width=0.8\textwidth]{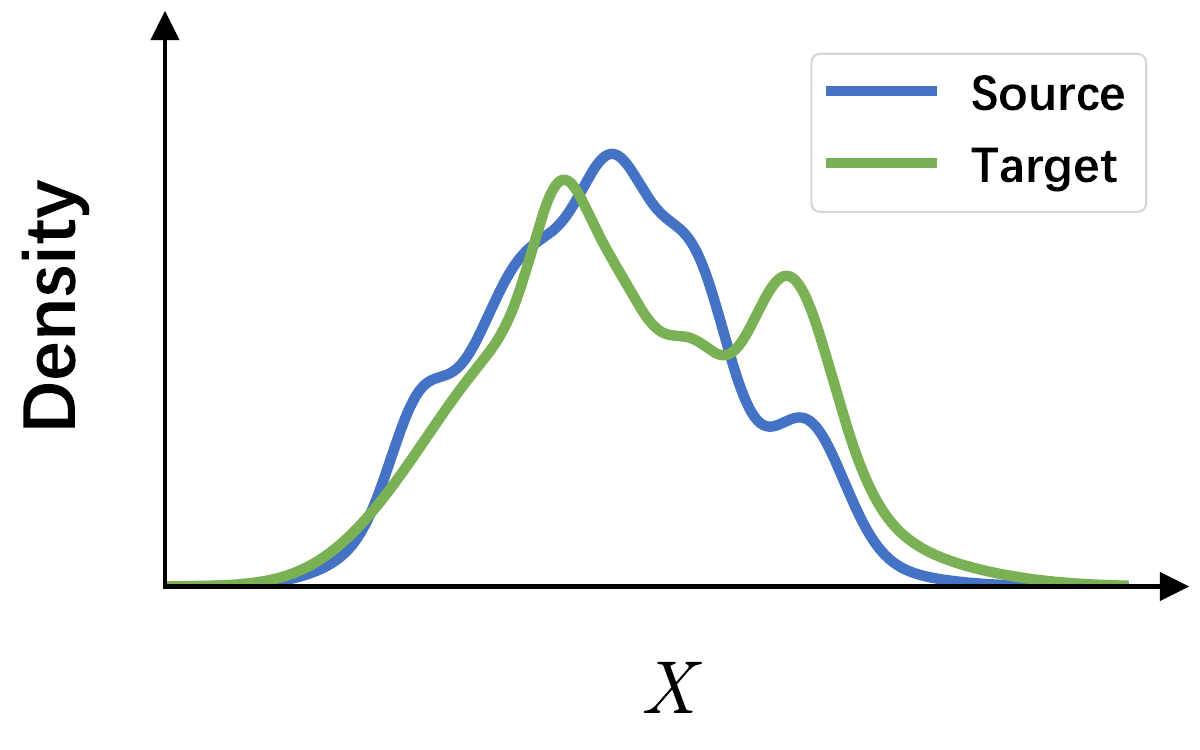}
    \caption{Individual Selection Bias}
  \end{subfigure}
  \quad
  \begin{subfigure}[t]{0.35\textwidth}
    \centering
    \includegraphics[width=0.8\textwidth]{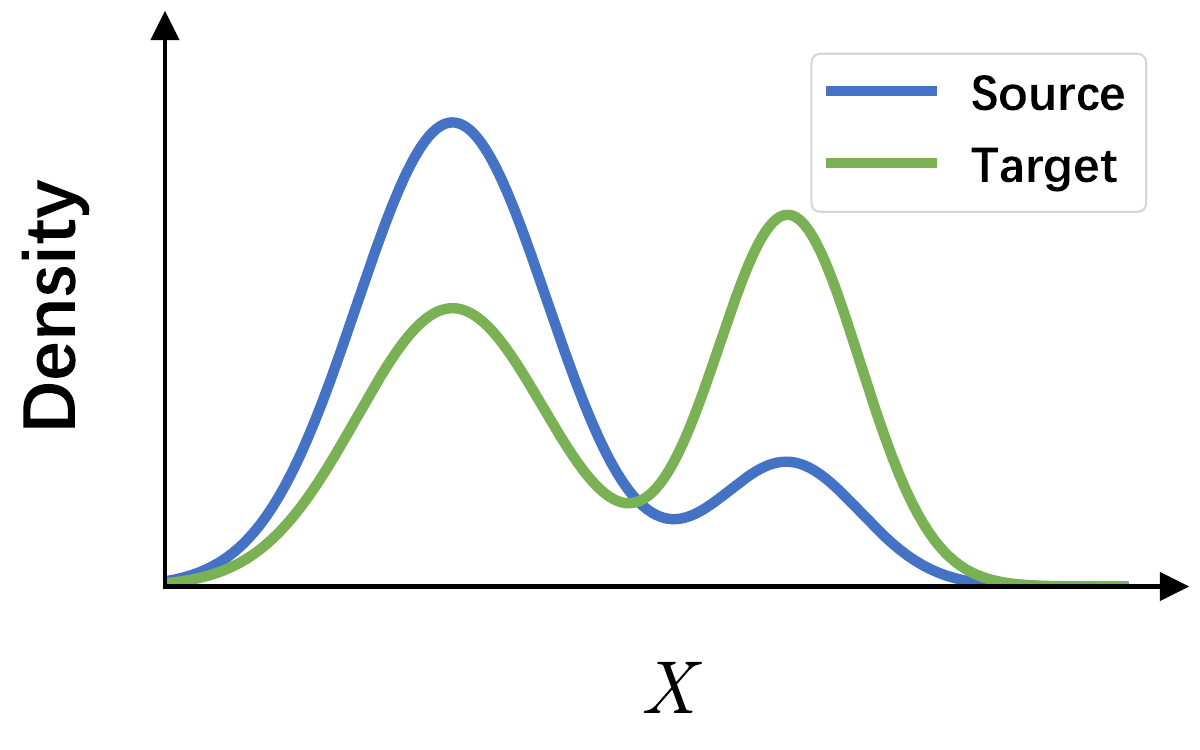}
    \caption{Group Selection Bias}
  \end{subfigure}
  \caption{An illustration of individual selection bias and group selection bias. The blue curves represent the input distribution of the source domain, and the green curves represent the input distribution of the target domain. For individual selection bias, the two distributions are known to differ. However, it is hard to attribute this difference to a certain subpopulation in the data. In contrast, for group selection bias, apparent changes in different subpopulations cause the distribution shift.}

  \label{fig:selection-bias}
\end{figure}

One type of selection bias is individual selection bias. We use the word ``individual'' to emphasize its difference from group selection bias (Figure \ref{fig:selection-bias}). Individual selection bias often appears for reasons such as variations in dataset re-collection~\citep{recht2019imagenet}, evolving clinical conditions~\citep{sahiner2023data}, or shifts in image styles~\citep{geirhos2018imagenettrained}. Additionally, researchers can simulate individual selection bias through controlled modifications like image coloring or corruption~\citep{nam2020learning}, distortion~\citep{qin2023effective}, and patch operations~\citep{qin2022understanding}. In all these cases that we consider as individual selection bias, the selection bias is applied individually to each data point. In contrast, in group selection bias, the selection bias is applied according to the group.

We use a definition of individual selection bias similar to Shimodaira's \citep{shimodaira2000improving}:
\begin{definition}[Individual Selection Bias]
\label{def:individual selection bias}
    Consider the prediction task of inferring $Y$ from $X$ under the source domain $e_s$ and the target domain $e_t$. Individual selection bias causes a distribution shift where
\[
P(X\mid E = e_s)\neq P(X \mid E = e_t),\ P(Y\mid X, E = e_s) = P(Y\mid X, E = e_t).
\]
\end{definition}
Figure \hyperlink{selection-bias}{3(a)} illustrates individual selection bias. When individual selection bias exists, input distributions of the source (blue curve) and target (green curve) domains differ.

Addressing individual selection bias is essential to mitigate distribution shifts that lead to performance drops during model testing. Methods designed to address this selection bias not only improve model performance but also intersect with broader AI safety issues, particularly the principles of AI democracy. In the following discussion, we explore the connection between methods for addressing individual selection bias and strategies for promoting AI democracy.

\subsubsection{Democracy}
\label{sec:democracy}

In this paper, AI democracy denotes the principle of broadening access to AI research~\citep{mushkani2025position, seger2023democratising}. With the rapid growth in the sizes of training datasets and models, researchers with limited computational resources often find themselves unable to develop or analyze state-of-the-art systems. This barrier not only restricts innovation but also limits the comprehensive evaluation of AI safety, as fewer experts are able to investigate the safety implications of these systems. Thus, AI democracy improves AI safety by allowing diverse researchers to scrutinize AI systems.

Data pruning is a promising method to democratize model training. By carefully selecting a subset of the full dataset, this method maintains a balance between the complexity and representativeness of examples, ensuring that models trained on the pruned data can perform on par with those trained on the entire dataset. The reduced dataset size lowers computational demands, which enables researchers with limited resources to train and evaluate advanced models~\citep{tan2024data}.

Data pruning can be seen as a dual problem of individual selection bias. To clarify this idea, we first formally define the data pruning task as follows~\citep{tan2024data}:

\begin{definition}[Data Pruning]
\label{def:Data Pruning}
    Given a full training set $D_\text{original}$ sampled from the original data distribution $P_{\text{original}}(X, Y)$, 
    data pruning aims to design a (stochastic) selection strategy $S:x\rightarrow \{0,1\}, x\sim P_{\text{original}}(X)$ with the following goal: The selected dataset $D_{\text{selected}}=\{x\in D_{\text{original}}\mid S(x)=1\}$ can be viewed as sampled from another distribution $P_{\text{selected}}(X, Y)$ that depends on the pruning strategy $S$ and the original distribution $P_{\text{original}}(X, Y)$. By training a model $f$ on the subset $D_{\text{selected}}$, the goal is to improve the performance of $f$ on $P_{\text{original}}(X, Y)$.
\end{definition}

\begin{tcolorbox}[title=Definition-Level Connection, colback=white, colframe=gray, fonttitle=\bfseries, breakable]
Both \textbf{individual selection bias} and \textbf{data pruning} induce a shift in the marginal distribution of $X$. In both cases, the goal is to train a model on one distribution (the source distribution in individual selection bias, or the selected data distribution in data pruning) such that it performs well on another distribution (the target distribution in individual selection bias, or the original data distribution in data pruning).
\end{tcolorbox}

From Definition \ref{def:individual selection bias} and Definition \ref{def:Data Pruning}, there is a natural correspondence between addressing individual selection bias and data pruning if we interpret $P_{\text{original}}(X,Y)$ as the target distribution and $P_{\text{selected}}(X,Y)$ as the source distribution. First, data pruning induces an analogous change in the marginal distribution of $X$ through its selection mechanism \(P_{\text{selected}}(X)
=
P_{\text{original}}(X\mid S=1)
\neq
P_{\text{original}}(X)\).

\begin{wrapfigure}[19]{r}{0.4\textwidth}
  \centering
  \vspace{-1.7em}
  \includegraphics[width=0.4\textwidth]{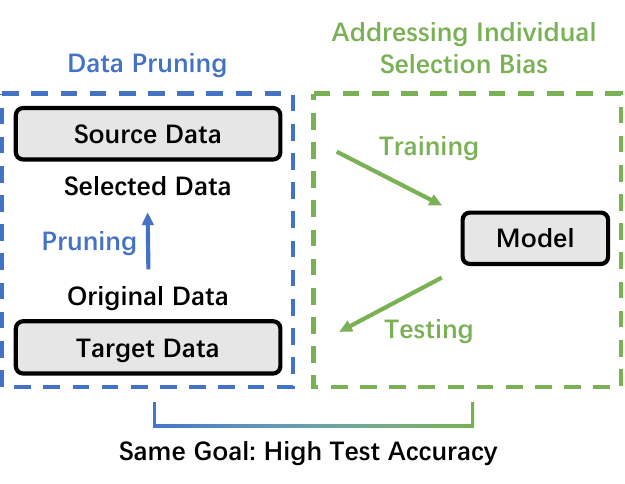} 
  \caption{Data pruning is a dual problem of addressing individual selection bias. To address individual selection bias, methods focus on refining the training process so that a model trained on source data performs well on target data. By comparison, data pruning methods prune less informative data from the original dataset, so that a model trained on the remaining subset still performs well. Both problems share the same goal of improving test accuracy under distribution differences.}
  \label{fig:dual-problem}
\end{wrapfigure}

Second, as illustrated in Figure \ref{fig:dual-problem}, techniques for addressing individual selection bias focus on refining the training process so that a model trained on source data performs well on target data. In contrast, data pruning methods select a representative subset from the full dataset, ensuring that a model trained on this subset achieves performance comparable to one trained on the entire dataset.  In both cases, the objective is to bridge the performance gap caused by differing input distributions between training and evaluation. Therefore, data pruning can be viewed as a dual problem of addressing individual selection bias.

Below, we compare techniques for addressing individual selection bias with those used in data pruning. Owing to the duality in these problems and the similarities in their underlying approaches, we explore how methods developed for one research direction can be adapted and applied interchangeably to the other.

To improve performance under individual selection bias, methods generally assign different weights to examples in the training set. Importance reweighting~\citep{shimodaira2000improving, fang2020rethinking} is a category of methods that mitigate the effect of individual selection bias. The basic importance reweighting method uses weights derived from the distributions $P(X, Y \mid E = e_s)$ and $P(X, Y \mid E = e_t)$. Instead of minimizing the empirical risk $\hat{\mathcal{R}}_{e_s}(f) = \frac{1}{n}\sum_{i=1}^{n} l\bigl(f(x^{(i)}_{e_s}),y^{(i)}_{e_s}\bigr)$, with importance reweighting, we minimize the weighted risk
\[
\begin{aligned}
&\frac{1}{n}\sum_{i=1}^{n} \frac{P(X = x^{(i)}_{e_s}, Y = y^{(i)}_{e_s} \mid E = e_t)}
{P(X = x^{(i)}_{e_s}, Y = y^{(i)}_{e_s} \mid E = e_s)}
l\bigl(f(x^{(i)}_{e_s}), y^{(i)}_{e_s}\bigr),
\end{aligned}
\]
which is an unbiased estimator of the target risk \(\mathcal{R}_{e_t}(f).\) Intuitively, when the probability density around a data point is larger in the target domain than in the source domain, the weight of the data point is increased, so that the model is encouraged to minimize the loss on this data point.

When dealing with data from complex modalities, deriving an exact analytical form of the distributions is often intractable. In such cases, importance reweighting methods instead use weights from other sources. For example, the Learning-from-Failure (LfF) method~\citep{nam2020learning} trains two models simultaneously. The first model, optimized with a generalized cross entropy loss, primarily learns from easier examples, thus performing poorly on more challenging ones. These hard examples are then assigned higher weights during the training of the second model. Similarly, the Just-Train-Twice (JTT) approach~\citep{liu2021just} divides the training process into two stages. In the first stage, a low-complexity model is trained using empirical risk minimization (ERM), which typically misclassifies challenging inputs. In the second stage, these misclassified examples are upweighted to train the final, more robust classifier.

Beyond importance reweighting, many methods for addressing individual selection bias fall under the framework of distributionally robust optimization (DRO)~\citep{ben2013robust}. With DRO, we minimize the risk
$$\sup_{q \in \mathcal{Q}}\hat{\mathcal{R}}_q(f) = \sup_{q \in \mathcal{Q}}\frac{1}{n}\sum_{i=1}^{n} l\bigl(f(x_q^{(i)}),y_q^{(i)}\bigr),$$ where $\mathcal{Q}$ is a set of distributions called the uncertainty set, and $(x_q^{(i)}, y_q^{(i)})$ are drawn i.i.d. from $q$. Depending on the choice of $\mathcal{Q}$, various DRO methods have been proposed, including Conditional-Value-at-Risk (CVaR) DRO~\citep{levy2020large}, $\chi^2$-DRO~\citep{hashimoto2018fairness}, and other DRO variants~\citep{zhai2021doro, yu2022fast}. 

In data pruning, each training data point is assigned a score using certain metrics, and those data points falling within a selected score range are chosen to form a small yet representative subset. Researchers have proposed a variety of metrics for scoring, such as representations~\citep{maharana2024mathbbd}, probability vectors~\citep{ben2024distilling, caldeira2025diffprob}, predicted labels~\citep{meding2022trivial}, confidence scores~\citep{Coleman2020Selection,cho2025lightweight}, and gradients~\citep{tan2024data}. Each of these metrics captures distinct aspects of the data’s importance or complexity, enabling a selection that preserves the essential characteristics of the full dataset while greatly reducing the computational load.

\begin{tcolorbox}[title=Future Directions, colback=white, colframe=c2, fonttitle=\bfseries, breakable]
Due to the dual nature of data pruning and addressing individual selection bias, methods developed for one can often be repurposed for the other. For instance, the metrics used to score data points in data pruning could serve as weights in importance reweighting techniques for addressing individual selection bias.

Conversely, weights obtained through importance reweighting can be used in data pruning strategies. Furthermore, the way DRO-based methods optimize over an uncertainty set can also be regarded as a process of assigning sample weights. Challenging examples receive higher weights, which can be used for data pruning.
\end{tcolorbox}

\subsection{Group Selection Bias}
\label{sec:group-selection-bias}

Group selection bias is a specific form of individual selection bias (see Figure \ref{fig:Relationship Distribution Shift}). Unlike biases that occur at the individual level, group selection bias arises when the composition of groups differs between the source and target domains. Although both domains may contain the same groups, the relative proportions of these groups vary. This phenomenon has been observed in various contexts, such as racial groups~\citep{karkkainen2021fairface} or different time periods~\citep{ni-etal-2019-justifying}. We define group selection bias and the distribution shift it causes as follows:

\begin{definition}[Group Selection Bias]
\label{def:group selection bias}
    Consider the prediction task of inferring $Y$ from $X$ under the source domain $e_s$ and the target domain $e_t$. Group selection bias causes a distribution shift where
\[
\begin{aligned}
    &\exists\, U_g\in\mathcal{G}:\quad P(X\in U_g \mid E = e_s) \neq P(X\in U_g \mid E = e_t),\\[1ex]
    &P(Y\mid X, E = e_s) = P(Y\mid X, E = e_t),
\end{aligned}
\]
    where \(U_g\) denotes the range of \(X\) corresponding to group \(g\), i.e., \(U_g \subset \mathcal{X}\) and \(\bigcup_{g\in\mathcal{G}} U_g = \mathcal{X}\). Here, \(\mathcal{G}\) represents the range of group labels \(G\). A common example is that a few majority groups dominate under $e_s$, while groups are uniformly distributed under $e_t$.
\end{definition}

Figure \hyperlink{selection-bias}{3(b)} illustrates group selection bias. When group selection bias exists, the most intuitive difference between the source and target domains lies in the varying group sizes (``Density'' in the figure) across the two domains.

\subsubsection{Fairness}
\label{sec:group-selection-bias-fairness}

Compared to the ideal case of perfect alignment between source and target distributions, group selection bias not only degrades overall model performance, but it also hurts group fairness. Figure \ref{fig:Fairness} shows different types of group fairness. In this section, we primarily consider group fairness defined in terms of the population risk of each group~\citep{williamson2019fairness}, which we refer to as risk parity. More notions of group fairness that are related to other causes of distribution shift will be introduced in Sections \ref{sec:environmental-change-fairness} and \ref{sec:environmental-change-fairness2}.

\begin{definition} [Risk Parity]
\label{def:risk parity}
    Let $X$ denote the covariates and let $Y$ denote the target variable. We say a predictive model $f$ satisfies risk parity with respect to the loss function \(l\) and attribute $R$ if, for all values $r_i, r_j$ in the range of $R$, $\mathcal{R}_{r_i}(f) = \mathcal{R}_{r_j}(f)$, i.e.,
\[
\mathbb{E}\big[\,l(f(X),Y) \mid R = r_i\big] = \mathbb{E}\big[\,l(f(X),Y) \mid R = r_j\big].
\]
\end{definition}

\begin{tcolorbox}[title=Definition-Level Connection, colback=white, colframe=gray, fonttitle=\bfseries, breakable]
When the group partition $\{U_g\}_{g\in\mathcal{G}}$ is determined by a protected attribute $R$, risk parity requires equal risks across groups. Under \textbf{group selection bias}, source ERM minimizes a group-weighted risk, which tends to overemphasize majority groups. This makes the learned model fail to achieve \textbf{risk parity}~\citep{Sagawa2020Distributionally}.
\end{tcolorbox}

When group selection bias exists between training and test domains, the model $f$ tends to focus on the majority groups while neglecting minority ones, leading to unequal accuracies of the groups~\citep{Sagawa2020Distributionally}. Considering the group $G$ that a data point belongs to as the protected attribute $R$, group selection bias naturally results in a model that does not satisfy risk parity.  Moreover, in the target domain, a more uniform distribution over groups amplifies the effect of low accuracy on minority groups, thereby degrading the model performance.

Since group selection bias is a specific instance of individual selection bias, many mitigation techniques overlap between the two. The key difference lies in the availability of group labels for group selection bias. With these labels, it is possible to explicitly enhance the influence of underrepresented groups during training. In the following sections, we illustrate how representative methods for addressing group selection bias help to achieve risk parity.

Importance reweighting~\citep{Sagawa2020Distributionally} is such a method, where the examples from each group are assigned a weight inversely proportional to the group size. Specifically, the objective of importance reweighting is to minimize the risk
$$\sum_{g\in\mathcal{G}}\hat{\mathcal{R}}_g(f) = \sum_{g\in\mathcal{G}}\frac{1}{n_g}\sum_{i=1}^{n_g}l(f(x_g^{(i)}),y_g^{(i)}),$$ where the examples $(x^{(i)}_g, y^{(i)}_g)$ are drawn i.i.d from distribution $P(X,Y \mid G=g)$, and $n_g$ is the number of examples drawn.
By assigning a larger weight to the minority group due to its smaller size, importance reweighting encourages the model to update in a way that treats all groups equally, thus promoting risk parity. Similarly, applying subsampling techniques to the majority group~\citep{idrissi2022simple} effectively reduces the classification loss of the minority group during training.

Another popular approach for addressing group selection bias is Group DRO~\citep{Sagawa2020Distributionally}. In Group DRO, the model is trained to minimize the highest empirical risk across all groups, that is
$$\max_{g\in\mathcal{G}}\hat{\mathcal{R}}_g(f) = \max_{g\in\mathcal{G}}\frac{1}{n_g}\sum_{i=1}^{n_g}l(f(x_g^{(i)}),y_g^{(i)}).$$
Under this worst-case risk framework, the model is optimized to minimize the risk of the group with the highest risk, ultimately aiming for uniform risk across all groups. To further ensure that the model optimizes to reduce loss on the groups with larger generalization gaps, the adjusted group DRO approach~\citep{Sagawa2020Distributionally} adds an extra loss term that is inversely proportional to the square root of the group’s size. 
In a similar vein, \citet{krueger2021out} introduce risk extrapolation (REx), where Minimax REx further penalizes the group with the highest risk and Variance REx penalizes the variance of group risks, both encouraging more uniform risks across groups under larger distribution shifts. Recent variants of Group DRO further address several limitations of the standard formulation. To address the issue that Group DRO requires predefined groups, adversarial group discovery methods have been proposed to discover error-prone groups and optimize the worst-group loss~\citep{paranjape2022agro}. Probabilistic Group DRO extends Group DRO by removing the requirement that each example has a deterministic hard group label~\citep{ghosal2023distributionally}. Moreover, recent work points out that traditional Group DRO can suffer from within-group distribution shift due to minority groups with limited samples, and proposes Hierarchical Group DRO to jointly handle inter-group and intra-group uncertainties~\citep{jo2025mitigating}.

\begin{tcolorbox}[title=Future Directions, colback=white, colframe=c2, fonttitle=\bfseries, breakable]
Even though various approaches have aimed to address group selection bias, empirical evidence shows that worst-group accuracy frequently falls below the average accuracy~\citep{Sagawa2020Distributionally}, suggesting that true risk parity is not fully realized at convergence. Motivated by the fundamental definitional connection, a promising future direction is to further develop optimization procedures that can more reliably approach risk parity in practice.
\end{tcolorbox}

\section{Spurious Correlation}
\label{sec:spurious-correlation}
Different from selection bias which focuses on distribution shifts caused by the sample selection process of input $X$, some types of distribution shift are closely related to the emergence of spurious correlations among different features in the training data.

Spurious correlation is the statistical relationship where two or more variables are associated but not causally related~\citep{simon1954spurious}. In parallel with the rapid development of ML techniques in recent years, the sensitivity of ML models to spurious correlations has been frequently observed in various fields, such as computer vision~\citep{izmailov2022feature, phan2024controllable}, natural language processing~\citep{veitch2021counterfactual, wang2021identifying}, and reinforcement learning (RL)~\citep{lyle2021resolving, cunha2025unifying}.

Consider the classification of waterbirds and landbirds in images as an example. If almost all waterbirds appear with water backgrounds and landbirds with land backgrounds during training, the ML classifier might learn to predict bird types based on the image's background due to inductive bias~\citep{yang2024identifying}. However, in the test environment, waterbirds may be found in land backgrounds, which could significantly reduce the model's classification accuracy. Therefore, the spurious correlation between the background $Z$ and the label $Y$, i.e., their non-causal relationship, leads to inconsistency in the relationship between $Z$ and $Y$ across the training and testing environments. This inconsistency results in poor generalization performance of the ML model.

From the example above, we observe that the spurious correlation can generally be regarded as arising from group selection bias. This is because, from the training to test environments, there is a change in the distribution of groups formed by different kinds of birds and their corresponding backgrounds. Specifically, under group selection bias (see Figure \ref{fig:Relationship Distribution Shift}), there are two causes directly associated with the occurrence of spurious correlations, namely, environmental change and content change. In Sections \ref{sec:environmental-change} and \ref{subsec:Content change}, environmental change and content change will be introduced, respectively. We will discuss how environmental change and content change are closely connected to spurious correlations, and further demonstrate they are inherently and naturally connected to different aspects of AI safety, such as fairness, trustworthiness, and security.

\subsection{Environmental Change}
\label{sec:environmental-change}
Distribution shifts occur when there are changes in the environment between training and testing. In tasks where the input variable $X$ is used to predict the label $Y$, when there is an environmental change, some features of \(X\) experience a change in their relationship with \(Y\) 
while others maintain their relationship with \(Y\).
Specifically, environmental change is defined as follows:

\begin{definition}[Environmental Change]
\label{def:Environmental Change}
Consider the prediction task of inferring $Y$ from $X$ under the environment variable \(E\) with possible values \(\mathcal{E}\). Denote the Bayes error~\citep{fukunaga2013introduction} for predicting $Y$ from $X$ in environment \(e \in \mathcal{E}\) as \(\epsilon_e(X)\), and define the set of feature extractors that (approximately) preserve the Bayes error in environment \(e\) as
\(
\mathcal{F}_e = \big\{\Phi:\; \epsilon_e\big(\Phi(X)\big) \approx \epsilon_e(X)\big\}.
\)
Then, environmental change with respect to environment variable \(E\) causes a type of distribution shift where there exist a content (invariant) feature $C(X) \in \bigcap_{e \in \mathcal{E}} \mathcal{F}_e$ and an environmental (spurious) feature $Z(X)$, such that
\[
\begin{aligned}
    &\exists\, e_{i_0}, e_{j_0}\in\mathcal{E}:\quad P\big(Y \mid Z(X), E=e_{i_0}\big) \neq P\big(Y \mid Z(X), E=e_{j_0}\big),\\[1ex]
    &\forall\, e_i, e_j\in\mathcal{E}:\quad P\big(Y \mid C(X), E=e_i\big) = P\big(Y \mid C(X), E=e_j\big).
\end{aligned}
\]
In concise conditional independence notation,
\(Y \not\perp\!\!\!\perp E \mid Z(X)\) and \(Y \perp\!\!\!\perp E \mid C(X)\).
\end{definition}

\begin{figure}[htbp]
  \centering
  \includegraphics[width=0.59\textwidth]{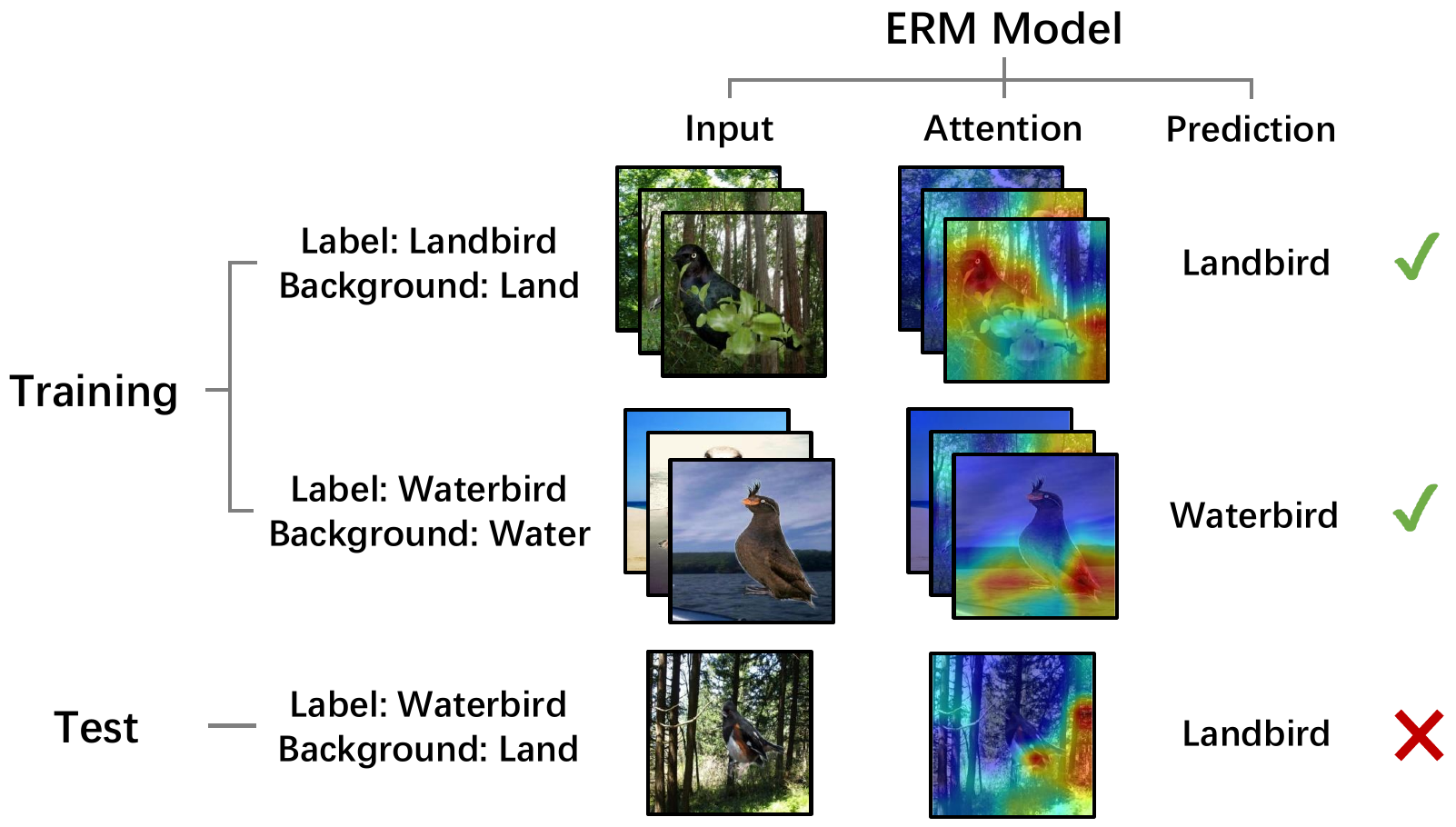} 
  \vspace{-0.2em}
  \caption{Illustration of the negative impact of environmental change on a model trained via empirical risk minimization (ERM), using the example of waterbird and landbird classification. The first column shows sample images from the training and test environments. In the training environments, most waterbirds have water backgrounds, and most landbirds have land backgrounds. However, in the test environment, the background is not highly correlated with the bird category; for instance, a waterbird can have a land background. Thus, $Z(X)$ represents the background feature, while $C(X)$ represents the bird feature. As shown in the second column with the Grad-CAM visualization~\citep{selvaraju2017grad} of the ERM model, the feature extractor $\Phi$ may focus on backgrounds $Z(X)$ to predict $Y$. Consequently, in the third column, the model's predicted labels indicate that the test model may misclassify a waterbird with a land background as a landbird based on $Z(X)$.}
  \label{fig:Environmental change}
\end{figure}

From Definition \ref{def:Environmental Change}, the inconsistent relationship between \( Z(X) \) and \( Y \) indicates the presence of environment-sensitive spurious correlation between them. Many models that predict label \( Y \) from \( X \) employ a function \( \Phi \) to extract features \(\Phi(X)\), which are then used for prediction. 
If the feature extractor \( \Phi \) captures information from environmental features \( Z(X) \), the model will inevitably learn these spurious correlations and suffer performance degradation under environmental changes. Figure \ref{fig:Environmental change} specifically illustrates how environmental changes occur in the previous example of waterbird versus landbird classification, and how the spurious correlation under environmental change affects the model's generalization performance.

Various methods have been employed to address the challenge of environmental change~\citep{arjovsky2019invariant, cai2019learning, koyama2020invariance, ahmed2020systematic, kirichenko2022last, wang2023disentangled, liu2025superclass}. Notably, many of these methods are closely related to several key aspects in AI safety, such as fairness, trustworthiness, and security. We will systematically analyze how the inherent similarities between environmental change and AI safety principles help bridge their methodologies.

\subsubsection{Fairness}
\label{sec:environmental-change-fairness}
From the perspective of the independence among random variables, group fairness mainly includes sufficiency, independence, and separation~\citep{raz2021group}. Specifically, for a model that uses $X$ to predict $Y$, let $\hat{Y}$ denote the model's prediction and $R$ denote the protected attribute. We define sufficiency, independence, and separation as follows:

\begin{definition}[Sufficiency, Independence, and Separation]
\label{def:SIS}
Let $\hat{Y}$ be a prediction derived from covariates $X$, let $Y$ denote the true outcome, and let $R$ denote the protected attribute. Then we have:
\[
\text{Sufficiency:}  \quad Y \perp\!\!\!\perp R \mid \hat{Y}, \quad
\text{Independence:}  \quad \hat{Y} \perp\!\!\!\perp R, \quad
\text{Separation:}  \quad \hat{Y} \perp\!\!\!\perp R \mid Y.
\]
\end{definition}

\begin{figure}[htbp]
  \centering
  \includegraphics[width=0.9\textwidth]{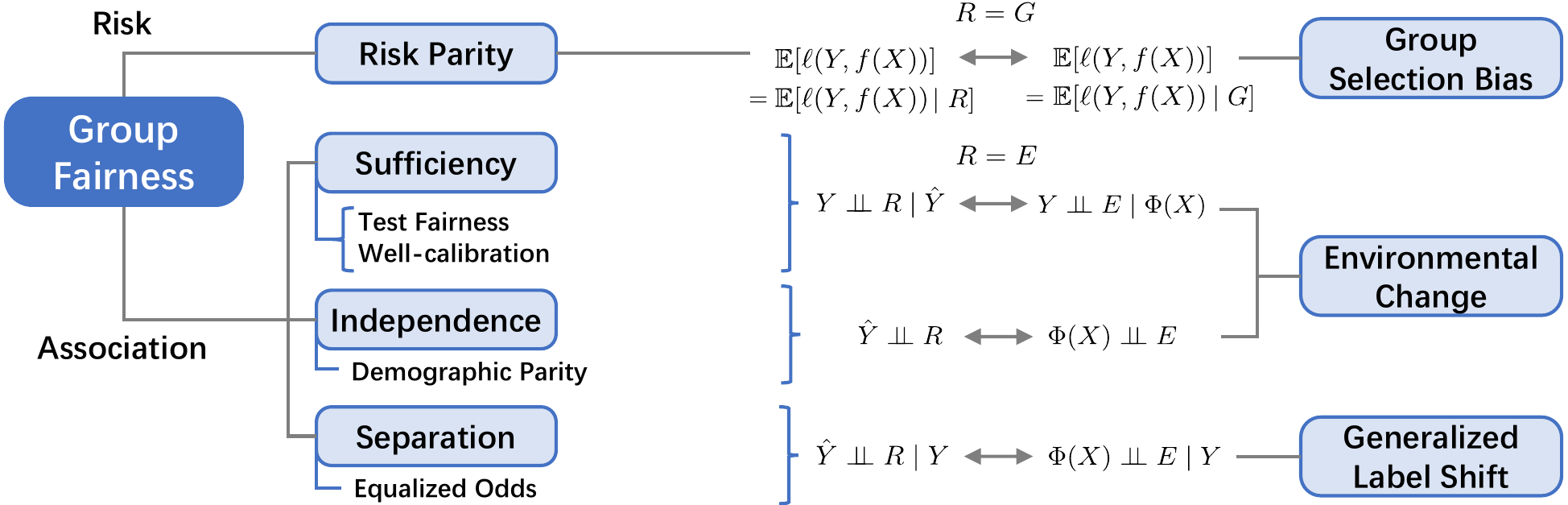} 
  \caption{Summary of group fairness and their relationships with different causes of distribution shift. For risk-based fairness notions, when considering the protected attribute $R$ as groups $G$ in distribution shift, risk parity (Definition \ref{def:risk parity}) and group selection bias (Definition \ref{def:group selection bias}) are closely related. From the perspective of the association between random variables, if $R$ is considered as the environmental variable $E$, then sufficiency and independence (Definition \ref{def:SIS}) are closely related to environmental change (Definition \ref{def:Environmental Change}), while separation (Definition \ref{def:SIS}) and generalized label shift (Definition \ref{def:Generalized Label Shift}) are closely related. Detailed relationships between different fairness definitions and the causes of distribution shift will be analyzed in the corresponding sections (Sections \ref{sec:group-selection-bias-fairness}, \ref{sec:environmental-change-fairness}, and \ref{sec:environmental-change-fairness2}).}

  \label{fig:Fairness}
\end{figure}

In Sections \ref{sec:environmental-change-fairness} and \ref{sec:environmental-change-fairness2}, we will provide a detailed analysis of the relationships between different types of distribution shifts and these three notions of group fairness. For clarity, we first present a unified list of the specific fairness definitions involved. Among these definitions, test fairness (Definition \ref{def:TestFairness}; see also \cite{chouldechova2017fair}) and well-calibration (Definition \ref{def:WellCalibration}; see also \cite{verma2018fairness}) belong to the group sufficiency notion, demographic parity (Definition \ref{def:DemographicParity}; see also \cite{mehrabi2021survey}) falls under group independence, and equalized odds (Definition \ref{def:EqualizedOdds}; see also \cite{hardt2016equality}) corresponds to group separation. Unlike the approaches surveyed by \citet{shao2024supervised}, which treat fairness with respect to a certain protected attribute as an additional constraint while attempting to maintain accuracy under distribution shift, we demonstrate that many methods for addressing distribution shift are inherently related to these fairness notions.

\begin{definition}[Test Fairness]
\label{def:TestFairness}
Let $S = S(x)$ be the predictive score computed from the covariates $X=x$, and let $Y \in \{0,1\}$ denote the true outcome indicator. Then, $S$ is test-fair with respect to the attribute $R \in \{r_1, r_2\}$ if for all values $s$ of $S$,
\(
P\big(Y = 1 \mid S = s, R = r_1\big) = P\big(Y = 1 \mid S = s, R = r_2\big).
\)
\end{definition}

\begin{definition}[Well-calibration]
\label{def:WellCalibration}
Let $S = S(x)$ be the predictive score computed from the covariates $X=x$, and let $Y \in \{0,1\}$ denote the true outcome indicator. Then, $S$ is well-calibrated with respect to the attribute $R \in \{r_1, r_2\}$ if for all values $s$ of $S$,
\(
P\big( Y = 1 \mid S = s, R = r_1\big) = P\big( Y = 1 \mid S = s, R = r_2\big) = s.
\)
\end{definition}

\begin{definition}[Demographic Parity]
\label{def:DemographicParity}
Let $\hat{Y} \in \{0,1\}$ be a prediction computed from the covariates $X$. Then, $\hat{Y}$ satisfies demographic parity with respect to the attribute $R \in \{r_1, r_2\}$ if
\(
P\big(\hat{Y} = 1 \mid R = r_1\big) = P\big(\hat{Y} = 1 \mid R = r_2\big).
\)
\end{definition}

\begin{definition}[Equalized Odds]
\label{def:EqualizedOdds}
Let $\hat{Y} \in \{0,1\}$ be a prediction computed from the covariates $X$, and let $Y \in \{0,1\}$ denote the true outcome indicator. Then, $\hat{Y}$ satisfies equalized odds with respect to the attribute $R \in \{r_1, r_2\}$ if for every $c \in \{0,1\}$,
\(
P\big(\hat{Y} = 1 \mid Y = c, R = r_1\big) = P\big(\hat{Y} = 1 \mid Y = c, R = r_2\big).
\)
\end{definition}

As shown in Figure \ref{fig:Fairness}, in this section, our primary focus is on the connection between environmental change and the sufficiency and independence conditions. Specifically, when environmental changes occur, it is crucial for models to capture appropriate features $\Phi(X)$ of $X$ for predicting $Y$. The Definition \ref{def:Environmental Change} suggests that models should avoid extracting environmental features $Z(X)$, and instead focus on capturing content features whose relationships with $Y$ remain invariant. This underpins the core idea of invariant learning.

Invariant learning aims to discover features $\Phi(X)$ that satisfy
\[
\forall\, e_i,e_j\in\mathcal{E}:\quad P\big(Y \mid \Phi(X), E=e_i\big) = P\big(Y \mid \Phi(X), E=e_j\big),
\]
where a classifier $\omega$ subsequently makes predictions based on $\Phi(X)$. The learned $\Phi$ avoids capturing spurious correlations that vary under environmental changes. 

\begin{tcolorbox}[title=Definition-Level Connection, colback=white, colframe=gray, fonttitle=\bfseries, breakable]
When the environment variable $E$ is interpreted as the protected attribute $R$ in group fairness, and the learned features $\Phi(X)$ or predictive score are interpreted as the score used in group fairness, robustness to spurious correlations under \textbf{environmental change} definitionally realizes \textbf{test fairness}.
When additional forms of invariance are imposed, the alignment extends to other fairness notions, including \textbf{well-calibration} and \textbf{demographic parity}.
\end{tcolorbox}

The definition of features $\Phi(X)$ is closely related to the notion of test fairness (Definition~\ref{def:TestFairness}), which is a specific case of group sufficiency~\citep{liu2019implicit}. In this setting, if we consider the environment variable $E$ as representing the protected attribute \(R\), then the representation $\Phi(X)$ in invariant learning is formally consistent with the predictive score $S(X)$ in test fairness, and can thus be regarded as an extension of sufficiency at the feature level. Moreover, the actual predictive score  $f(x) := (\omega \circ \Phi)(x)$ of the model satisfies the definition of test fairness when $\Phi(X)$ meets the minimal sufficiency condition. This means that $\Phi(X)$ is a sufficient statistic for $Y$ that is minimal in the sense that it contains no redundant or extraneous information.
 
Therefore, due to the mathematical connection between environmental change and test fairness, invariant learning techniques developed to address environmental change can be applied to achieve test fairness. Previous works implement or approximate invariant learning in different ways. 
While contributing to achieving test fairness, some of these methods are also related to other fairness notions.\footnote{Throughout this paper, when a method is associated with multiple fairness-related causes of distribution shift, we assign it to the section corresponding to the last cause of distribution shift.} 

\subsubsubsection{Conditional Distribution Invariance}
\label{sec:conditional-distribution-invariance}
To pursue invariant learning, we define conditional distribution invariance as the objective of methods that directly aim to keep the conditional probability \( P\big(Y \mid \Phi(X), E=e\big) \) unchanged across different \( e \).

\paragraph{Invariant Optimal Classifier Learning.
} When predicting $Y$ through $\Phi(X)$ in various environments, invariant optimal classifier learning requires the model to learn using the same optimal classifier $\omega$ in different environments. Notably, the invariance of $\omega$ has a close connection with learning $\Phi(X)$ such that $P\big(Y \mid \Phi(X), E=e\big)$ remains invariant across different environments $e$. \citet{peters2016causal} point out that when $\Phi(X)$ is a subset of $X$, invariant learning is equivalent to the existence of an invariant classifier $\omega$ across environments and an error distribution $\epsilon_e$, such that 
\(
Y = \omega\big(\Phi(X), \epsilon_e\big)\) and \(\epsilon_e \perp\!\!\!\perp \Phi(X)\)
for $X$ and $Y$ in all environments $e \in \mathcal{E}$. Based on this equivalence, Invariant causal prediction (ICP)~\citep{peters2016causal} estimates valid \(\Phi(X)\) using causal inference under the assumption that \(\omega\) is linear. \citet{heinze2018invariant} further extend the method of ICP to nonlinear settings.

Another common approach for invariant optimal classifier learning is to approximate the probability distribution in invariant learning using expectations. That is, for all the possible value $\alpha$, \[
\forall\, e_i,e_j\in\mathcal{E}:\quad \mathbb{E}\bigl[\, Y \,\big|\, \Phi(X)=\alpha, E=e_i \bigr]
=
\mathbb{E}\bigl[\, Y \,\big|\, \Phi(X)=\alpha, E=e_j \bigr]. 
\]
The representation $\Phi(X)$ is said to satisfy the environment invariance constraint (EIC)~\citep{creager2021environment} if it meets the above condition. EIC is a weaker constraint for invariant learning and can itself be interpreted as a fairness notion under group sufficiency~\citep{ creager2021environment}. Moreover, when $Y$ is binary, EIC is equivalent to invariant learning. Thus, it is also closely related to test fairness.

EIC is intimately connected to invariant optimal classifier learning because, for a given environment $e$ and feature $\Phi(x) = \alpha$, the Bayes classifier $\omega := \omega^*_{e}$ that minimizes the population risk $\mathcal{R}_{e}(\omega \circ \Phi) = \mathbb{E}\bigl[\,\ell\bigl((\omega \circ \Phi)(X), Y\bigr) \,\big|\, E=e\bigr]$ can be expressed as \(
(\omega^*_e \circ \Phi)(x)
=
\mathbb{E}\bigl[\,Y \,\big|\, \Phi(X)=\alpha,\; E=e \bigr]
\)
when the loss function $\ell$ is the mean squared error~\citep{rojas2018invariant, arjovsky2019invariant}. This shows that achieving EIC is equivalent to identifying an invariant optimal classifier $\omega$ across all environments.

Among various algorithms that implement EIC, Invariant Risk Minimization (IRM)~\citep{arjovsky2019invariant} is one of the most representative methods. Its core idea precisely lies in finding an invariant optimal classifier $\omega$ when using $\Phi(X)$ to predict $Y$ across different environments. In the subsequent work of IRM, Ensemble Invariant Risk Minimization Games (EIRM)~\citep{ahuja2020invariant} attempt to find the Nash equilibrium of an ensemble game among different environments and prove that EIRM satisfies EIC under appropriate conditions. \citet{teney2021unshuffling} minimize the risk across different environments while using the variance of the classifiers $\omega$ under different environments as a regularizer, so that the model is encouraged to reach a stationary point that satisfies the IRM principle. In Regret Minimization~\citep{jin2020domain}, the IRM constraint is substituted with a predictive regret formulation, which can be minimized if there exists a single optimal classifier under all environments. 

Furthermore, some works have extended IRM by incorporating additional constraints. \citet{ahuja2021invariance} argue that EIC alone is not sufficient to guarantee good generalization performance, and they introduce the information bottleneck constraint to address the issue that \(\Phi(X)\) might be fully informative about \(Y\). \citet{wald2021calibration} analyze the relationship between EIC and calibration of predictive functions. Building upon IRM, they require the model $f = \omega \circ \Phi$ be calibrated across all environments. That is, for every $\alpha$ in the range of $f$:
\[
\forall\, e_i,e_j\in\mathcal{E}:\quad
\mathbb{E}\bigl[\, Y \,\big|\, f(X)=\alpha,\; E=e_i \bigr]
=
\mathbb{E}\bigl[\, Y \,\big|\, f(X)=\alpha,\; E=e_j \bigr]
=
\alpha.
\]
This requirement is closely related to the fairness notion named well-calibration (Definition \ref{def:WellCalibration}), which is an extension of test fairness. 

\paragraph{Conditional Probability Matching.} Different from methods that learn an invariant optimal classifier and focus on the EIC, some models directly estimate $\Phi(X)$ by matching the conditional distributions $P\big(Y \mid \Phi(X), E = e\big)$ across different environments $e$. 

The Maximal Invariant Predictor (MIP)~\citep{koyama2020invariance} aligns the conditional distributions $P\big(Y \mid \Phi(X), E = e\big)$ and $P\big(Y \mid \Phi(X)\big)$ for each $e$ by minimizing their Kullback-Leibler (KL) divergence, while simultaneously maximizing the Shannon mutual information between $\Phi(X)$ and $Y$. Heterogeneous Risk Minimization (HRM)~\citep{liu2021heterogeneous} builds on MIP to address situations where environmental labels are unknown. \citet{muller2021learning} leverage conditional normalizing flows to identify an equivalent form of the conditional independence $Y \perp\!\!\!\perp E \mid \Phi(X)$. The Latent Causal Invariance Model (LaCIM)~\citep{sun2021recovering} directly employs a Variational Auto-Encoder to obtain invariant $P\big(Y \mid \Phi(X), E = e\big)$ across environments $e$, under the additional assumption of invariance in $P\big(X \mid \Phi(X), Z(X), E = e\big)$. 

Furthermore, a series of methods use mutual information techniques to achieve conditional probability matching. For instance, Invariant Rationalization~\citep{chang2020invariant} reformulates the conditional independence $Y \perp\!\!\!\perp E \mid \Phi(X)$ via Shannon entropy:
\(
H\big(Y \mid \Phi(X), E\big) = H\big(Y \mid \Phi(X)\big),
\)
i.e., the mutual information $I\big(Y; E \mid \Phi(X)\big)$ equals zero. At the same time, it maximizes the mutual information between $\Phi(X)$ and $Y$. Similarly, \citet{li2022invariant} propose an invariant information bottleneck principle based on minimizing $I\big(Y; E \mid \Phi(X)\big)$. 

When explicit environmental labels are unavailable, but knowledge of spurious features $Z(X)$ under environmental changes is given\footnote{Throughout this paper, ``knowledge of $Z(X)$'' refers to either direct access to $Z(X)$~\citep{puli2021out,puli2022nuisances, makar2022causally, oh2022learning} or reference models capturing $Z(X)$-related information~\citep{zheng2022causally, zhang2022correct}. We treat them uniformly since we focus on how the model uses $Z(X)$ (regardless of acquisition method) to analyze connections with ML fairness. Similarly, ``access to environment $E$'' is treated in the same way.}, some models instead learn $\Phi(X)$ such that $P\big(Y \mid \Phi(X), Z(X) = \alpha\big)$ remains consistent across $\alpha$\footnote{Unless otherwise specified, we define all notions of invariance with respect to the environment label \(E\). Whenever \(E\) is unavailable but the spurious feature \(Z(X)\) is known, we naturally refer to the same notion of invariance across different values of \(Z(X)\).}. \citet{puli2021out} propose finding a distribution where $Y \perp\!\!\!\perp Z(X)$. Then it learns $\Phi(X)$ by minimizing the mutual information between $Y$ and $Z(X)$ given $\Phi(X)$, such that $Y \perp\!\!\!\perp Z(X) \mid \Phi(X)$, while requiring that $\Phi(X)$ has maximal mutual information with $Y$. Since annotations for $Z(X)$ are seldom available, Nuisances via Negativa~\citep{puli2022nuisances} constructs semantic corruptions $T(X)$ of $X$ and demonstrates that it serves well as a proxy for $Z(X)$.

\subsubsubsection{Marginal Distribution Invariance}
\label{sec:marginal-distribution-invariance}
Marginal distribution invariance refers to the objective of learning $\Phi(X)$ such that the marginal distribution $P\big(\Phi(X) \mid E=e\big)$ remains invariant across different environments $e$. Under the additional condition that the invariance of $P\big(Y \mid \Phi(X), E=e\big)$ also holds, the joint distribution
\[
P\big(\Phi(X), Y \mid E=e\big) = P\big(Y \mid \Phi(X), E=e\big)\,P\big(\Phi(X) \mid E=e\big)
\]
is also invariant across different environments $e$, which implies that minimizing the population risk in different environments yields the same solution $\omega$. With the same prediction function $\omega$, the invariance of $P\big(\Phi(X) \mid E=e\big)$ also implies that the model $f = \omega \circ \Phi$ satisfies the invariance of $P\big(f(X)\mid E=e\big)$. This is equivalent to the group independence fairness definition known as demographic parity (Definition \ref{def:DemographicParity}).

To achieve invariant learning based on the invariant marginal distribution, various approaches directly assume or enforce the invariance of $P\bigl(Y \mid \Phi(X), E=e\bigr)$, and then further learn feature $\Phi(X)$ across different environments such that the discrepancy between them is minimized.

Specifically, Maximum Mean Discrepancy~\citep{borgwardt2006integrating} (MMD) is widely used as a tool for minimizing the distance between different marginal distributions. Early methods such as Maximum Mean Discrepancy Embedding~\citep{pan2008transfer}, Transfer Component Analysis~\citep{pan2010domain}, Domain Invariant Projection~\citep{baktashmotlagh2013unsupervised}, Transfer Joint Matching~\citep{long2014transfer}, and Deep Adaptation Networks~\citep{long2015learning} use MMD to ensure that $P\big(\Phi(X) \mid E=e\big)$ remains consistent across various environments. \citet{zellinger2017central, zellinger2019robust} propose the Central Moment Discrepancy as a computationally more efficient alternative to MMD. Domain-Adversarial Neural Networks~\citep{ganin2016domain} aim to minimize the $\mathcal{H}$-divergence between $\Phi(X)$ across different environments while ensuring good predictive performance in environments with known labels. \citet{shen2018wasserstein} use the Wasserstein distance for minimizing the distance between the distributions of $\Phi(X)$ from different environments. Since the pairwise distances between $P\big(\Phi(X), Y \mid E = e\big)$ for different $e$ can be difficult to optimize, \citet{chevalley2022invariant} introduce a softer constraint that offers a better trade-off between invariance and predictability.

In addition to requiring the invariance of marginal distributions and assuming the invariance of $P\bigl(Y \mid \Phi(X), E=e\bigr)$, some methods directly maintain the invariance of both $P\big(Y \mid \Phi(X), E=e\big)$ and $P\big(\Phi(X) \mid E=e\big)$ across different environments $e$. Domain Invariant Component Analysis~\citep{muandet2013domain} minimizes the distance between the distributions of $\Phi(X)$ across environments and ensures an invariant conditional relationship of $Y$ given $\Phi(X)$. \citet{zhao2020domain} achieve invariant marginal distributions through adversarial training, and incorporates entropy regularization to minimize the divergence between conditional distributions across different $e$. \citet{nguyen2021domain} simultaneously align these two distributions by leveraging Domain Density Transformation Functions. While enforcing the invariance of the marginal distribution, \citet{katz2024supervised} introduce the cross-domain conditionals agreement error to quantify the discrepancy of $P\big(Y \mid \Phi(X), E=e\big)$ across different environments.

Similar to the case when achieving conditional distribution invariance, there may also be a lack of environmental labels $E$, but spurious features $Z(X)$ under environmental changes can be obtained. In this case, \citet{makar2022causally} first find a distribution under which $Y \perp\!\!\!\perp Z(X)$. Then, based on this distribution, \citet{makar2022causally} instead learn $\Phi$ satisfying $\Phi(X) \perp\!\!\!\perp Z(X)$  by minimizing the MMD of $P\big(\Phi(X) \mid Z(X)\big)$ under different $Z(X)$. \citet{zheng2022causally} extend \citet{makar2022causally} to more general settings regarding the dimensionality and type of spurious features and labels.

\begin{tcolorbox}[title=Future Directions, colback=white, colframe=c2, fonttitle=\bfseries, breakable]
The close relationship between improving robustness under environmental change and achieving different group fairness objectives suggests that the methods reviewed above can substantially enrich the methodological toolbox for group fairness. 
Importantly, the potential application of these methods is not limited to cases where the protected attribute $R$ is exactly the environment variable $E$ that induces spurious correlations. They can also be adapted to broader group fairness settings involving general protected attributes $R$, such as race, gender, and age, which are more commonly considered in standard group fairness research~\citep{pessach2022review}.

\end{tcolorbox}

\subsubsection{Trustworthiness}
\label{sec:environmental-change-trustworthiness}

To ensure the safety of AI systems, many guidelines have been proposed for building trustworthy AI~\citep{kaur2022trustworthy,li2023trustworthy,herrera2025responsible}. Overall, AI trustworthiness requires that the model effectively encodes the user's goals to satisfy their expectations, and that its decision-making process can be understood and verified~\citep{isoiec_tr24028_2020}. Consequently, both alignment and interpretability are essential for achieving AI trustworthiness. In this section, we discuss how these two aspects of trustworthiness are related to environmental change from a definitional perspective.

\subsubsubsection{Interpretability}
\label{sec:environmental-change-interpretability}
Modern deep neural networks are often difficult to interpret because of their complex architectures and black-box behavior, making interpretability an important topic in AI safety~\citep{li2022interpretable}. Broadly, interpretability refers to explaining or presenting model behavior in human-understandable terms~\citep{doshi2017towards}. In this section, we focus on the interpretability of latent representations, especially disentangled representations. Following~\citet{wang2024disentangled}, a disentangled representation can be defined as follows:

\begin{definition}[Disentangled Representation]
\label{def:Disentangled Representation}
 Disentangled representation should separate the distinct, independent and informative generative factors of variation in the data. Single latent variables are sensitive to changes in single underlying generative factors, while being relatively invariant to changes in other factors.
\end{definition}

\begin{tcolorbox}[title=Definition-Level Connection, colback=white, colframe=gray, fonttitle=\bfseries, breakable]
Mitigating the impact of \textbf{environmental change} can be operationalized by identifying and separating $Z(X)$ from $C(X)$, which constitutes a concrete form of \textbf{representation disentanglement} between the environmental factor and the content factor.
\end{tcolorbox}

Many works have achieved the separation between $Z(X)$ and $C(X)$ by disentangling the features of $X$ based on their relationships with the target label $Y$ and the environment $E$. Note that different works may refer to the environment \(E\) as domain and the label \(Y\) as class, identity, or category, depending on the specific context. For example, the Disentangled Semantic Representation model~\citep{cai2019learning} employs gradient reversal layers to construct domain and label classifiers, which are used to partition the latent variables into semantic and domain components. Domain2Vec~\citep{peng2020domain2vec} achieves the disentanglement of domain-specific and category-specific features through Gram matrix. \citet{peng2019domain} decompose features into domain-invariant and domain-specific components while removing information from class-irrelevant features. MDTA-ITA~\citep{gholami2020unsupervised} uses two distinct encoders to capture domain-specific and shared information across different domains $e$. \citet{zhang2022towards2} separate semantic and variation factors of $X$ through solving a constrained optimization problem. Chroma-VAE~\citep{yang2022chroma} mitigates spurious correlations by constructing shortcut-invariant and shortcut-encoded latent spaces. \citet{wang2023disentangled} separate domain-related and domain-unrelated semantic features, and then refine the semantic features with causality. 

While all aforementioned methods incorporate autoencoder architectures and their variants, Disentangled Feature Augmentation~\citep{lee2021learning} leverages feature-level data augmentation to embed $X$ into intrinsic and bias attributes. Generalizable Person Re-Identification~\citep{zhang2022learning} extracts identity-specific and domain-specific features, and then employs backdoor adjustment to mitigate the influence of domain-specific features on identity-specific features.

In summary, these feature disentanglement methods that are investigated in this section typically partition the features of $X$ into two components to represent environmental features \(Z(X)\) and content features \(C(X)\) separately. After isolating these components, by predicting \(Y\) based on \(C(X)\), the models can effectively mitigate the impact of spurious correlations under environmental changes. In this manner, disentangled representation learning enables a more interpretable extraction and analysis of features. Moreover, we argue that, from another view of the cause of spurious correlation, the features of $X$ can be disentangled in a more structured and granular manner. Further details will be provided in Section \ref{sec:content-change-trustworthiness}.

\begin{tcolorbox}[title=Future Directions, colback=white, colframe=c2, fonttitle=\bfseries, breakable]
Separating environmental features \(Z(X)\) from content features \(C(X)\) provides a concrete way to achieve representation disentanglement for mitigating environmental change. Moreover, existing methods often support such separation through qualitative evidence (e.g., t-SNE visualization of feature alignment~\citep{cai2019learning}, Grad-CAM analyses~\citep{yang2022chroma}) and proxy metrics (e.g., out-of-distribution accuracy~\citep{lee2021learning, zhang2022towards2}). These evaluations suggest that the learned components are useful for mitigating environmental change, but they do not fully quantify whether \(Z(X)\) and \(C(X)\) are disentangled in the sense required by this setting. Therefore, an important future direction is to develop or adapt methods whose disentanglement properties can be systematically evaluated using more explicit disentanglement metrics tailored to environmental change.
\end{tcolorbox}

\subsubsubsection{Alignment}
\label{sec:environmental-change-alignment}
As society increasingly delegates decision-making to AI systems, a fundamental challenge surfaces: How can we ensure these technologies authentically reflect human ethics and societal norms? This question lies at the heart of AI alignment research, which seeks to ensure that AI systems behave in ways that are in accordance with human intentions and values~\citep{ji2023ai}, particularly in complex systems such as LLMs~\citep{shen2023large} and RL applications~\citep{kaufmann2023survey}. It is widely recognized that inner and outer alignment constitute two critical components of AI alignment research~\citep{shen2024towards}. In general, outer alignment refers to the alignment of human values with the AI system's training objectives, and inner alignment refers to the AI system's actual achievement to these training objectives during operation~\citep{hubinger2019risks}. In other words, outer alignment ensures that the training goals of AI systems are those we value as beneficial, while inner alignment ensures that the systems faithfully pursue these goals in practice. 

\begin{figure}[htbp]
  \centering
  \includegraphics[width=0.75\textwidth]{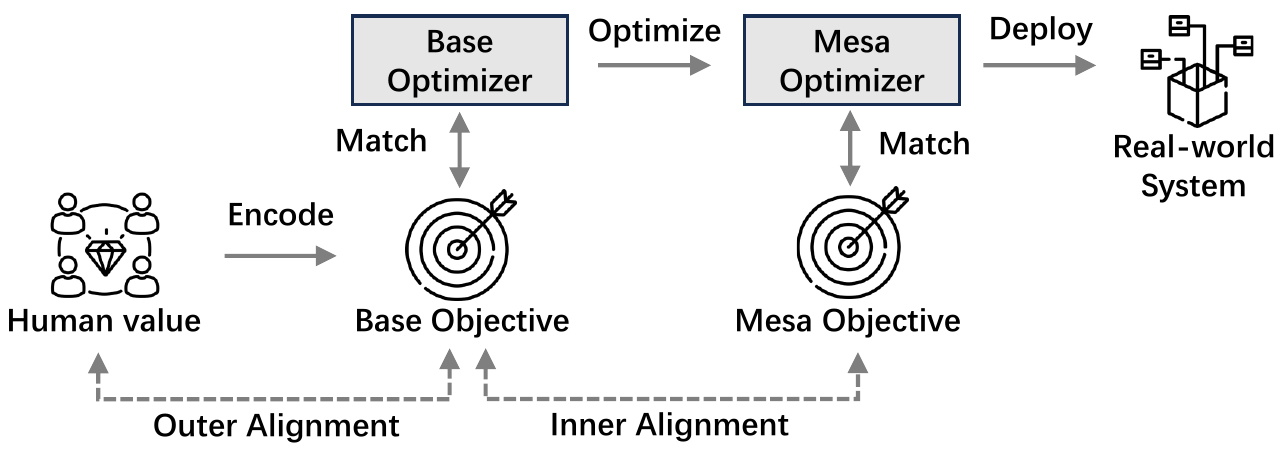} 
  \caption{The relationship between inner alignment and outer alignment.}
  \label{fig:Alignment}
\end{figure}

To provide a more precise definition of inner and outer alignment, \citet{hubinger2019risks} introduce four relevant concepts. The base objective is defined as the initial set of goals or tasks that the AI system is programmed to optimize, and the base optimizer is the algorithmic framework or mechanism through which these goals are pursued. The mesa-objective is the objective discovered by the base optimizer that shows effective performance in the training environment. Consequently, at this point, the base optimizer turns into the mesa-optimizer to achieve the mesa-objective. With these concepts, \citet{hubinger2019risks} define inner alignment and outer alignment as follows (see Figure \ref{fig:Alignment}):

\begin{definition}[Inner and Outer Alignment]
\label{def:Inner and Outer Alignment}
Inner Alignment: The alignment between the base objective and the mesa-objective.
Outer Alignment: The alignment between the base objective and the human values.
\end{definition}

Failure to achieve proper alignment in AI systems can lead to various detrimental outcomes, including behaviors that deviate from the principles of helpfulness, honesty, and harmlessness,
which together constitute the HHH alignment framework~\citep{askell2021general}. Goal misgeneralization, which is one of the numerous potential forms of misalignment, is considered to be caused by distribution shift~\citep{shah2022goal,di2022goal}. Specifically, we will focus on spurious correlations under environmental changes and reveal their close relationship with inner alignment. 

\begin{tcolorbox}[title=Definition-Level Connection, colback=white, colframe=gray, fonttitle=\bfseries, breakable]
Under environmental change, the presence of a spurious correlation between the environmental feature \(Z(X)\) and the label \(Y\) may lead the model to predict \(Y\) by relying on \(Z(X)\). If predicting \(Y\) from \(X\) is interpreted as the intended base objective and predicting \(Y\) through \(Z(X)\) as the learned mesa-objective, then the inconsistency in the relationship between \(Z(X)\) and \(Y\) under \textbf{environmental change} leads to a mismatch between the base objective and the mesa-objective, corresponding to \textbf{inner alignment} failure.
\end{tcolorbox}

Spurious correlation under environmental change can therefore be viewed as a direct trigger of inner alignment failure. Consequently, the approaches mentioned in Section \ref{sec:spurious-correlation}, which are designed for environmental change and directly connected to specific AI safety topics, as well as other efforts aimed at mitigating spurious correlations caused by environmental changes~\citep{kirichenko2022last, deng2023robust, wen2025elastic}, can contribute to improving the model's inner alignment.

Beyond the task of predicting \(Y\), in the realms of RL and LLMs, spurious correlation also serves as a major cause of inner alignment failure. For example, in the context of RL, if the agent designed to assist with driving is trained in an environment where traffic jams only occur during the daytime, it might learn a spurious correlation between light intensity and traffic congestion~\citep{ding2024seeing}. When deployed in a new environment where traffic jams occur at night, due to the misalignment between the mesa-objective and the base objective, the agent may fail to effectively predict or respond to traffic conditions. In the case of LLMs, models primarily trained on multiple-choice safety questions may associate safe answers with a particular answer style. This spurious correlation can result in the models failing to generate safe outputs when facing open-ended queries~\citep{wang2023fake}. Furthermore, with the emergence of advanced AI systems, deceptive alignment has been observed~\citep{hubinger2024sleeper}. In deceptive alignment, AI models' mesa-objectives are aware of the existence of the base objective and engage in deceptive behavior, pretending to align with the base objective to avoid being modified during training. However, once deployed in a new environment, the model will no longer follow the base objective and may exhibit undesirable behaviors. 

\begin{tcolorbox}[title=Future Directions, colback=white, colframe=c2, fonttitle=\bfseries, breakable]
The definition-level connection above suggests that, in prediction tasks, methods for mitigating spurious correlations under environmental change can contribute to inner alignment. Moreover, since inner alignment is also commonly discussed in more advanced AI systems, such as RL agents and LLMs, this connection further suggests the potential to apply these methods to more complex settings. A small but growing body of work has begun to adapt concepts from environmental changes to RL and LLMs, including invariant learning~\citep{sonar2021invariant, saengkyongam2023invariant, zheng2023improving} and causal representation learning~\citep{zhang2025causal}. Meanwhile, research on deceptive alignment remains largely focused on conceptual analysis, empirical evaluation, and early-stage mitigation~\citep{shen2023large,carranza2023deceptive,hubinger2024sleeper, ji2025mitigating}. This leaves substantial room for future work to transfer techniques for addressing environmental change in prediction tasks to inner alignment challenges in broader settings.
\end{tcolorbox}

\subsubsection{Security}
\label{sec:environmental-change-security}

While ML models have achieved remarkable success across various fields, they also face significant security challenges. ML attacks refer to malicious attempts to subvert ML systems through adversarial inputs~\citep{pauling2022tutorial, hua2024initialization}. These attacks enable adversaries to compromise model integrity, availability, and privacy~\citep{biggio2018wild}. Specifically, backdoor attacks~\citep{gu2017badnets} constitute a critical focus within ML attack research, posing substantial risks to computer vision~\citep{zhao2020clean}, natural language processing~\citep{yang2021rethinking}, and RL applications~\citep{ wang2021backdoorl}.

Backdoor attacks aim to manipulate a model's behavior by introducing carefully crafted malicious patterns during training. These malicious patterns cause the model to exhibit unexpected behaviors when they are present during inference. While backdoors can be injected at various attack surfaces of the ML pipeline, including data, model, and code~\citep{gao2020backdoor}, this section primarily focuses on data poisoning-based backdoor attacks during the model training process. 

\begin{figure}[htbp]
  \centering
  \includegraphics[width=0.75\textwidth]{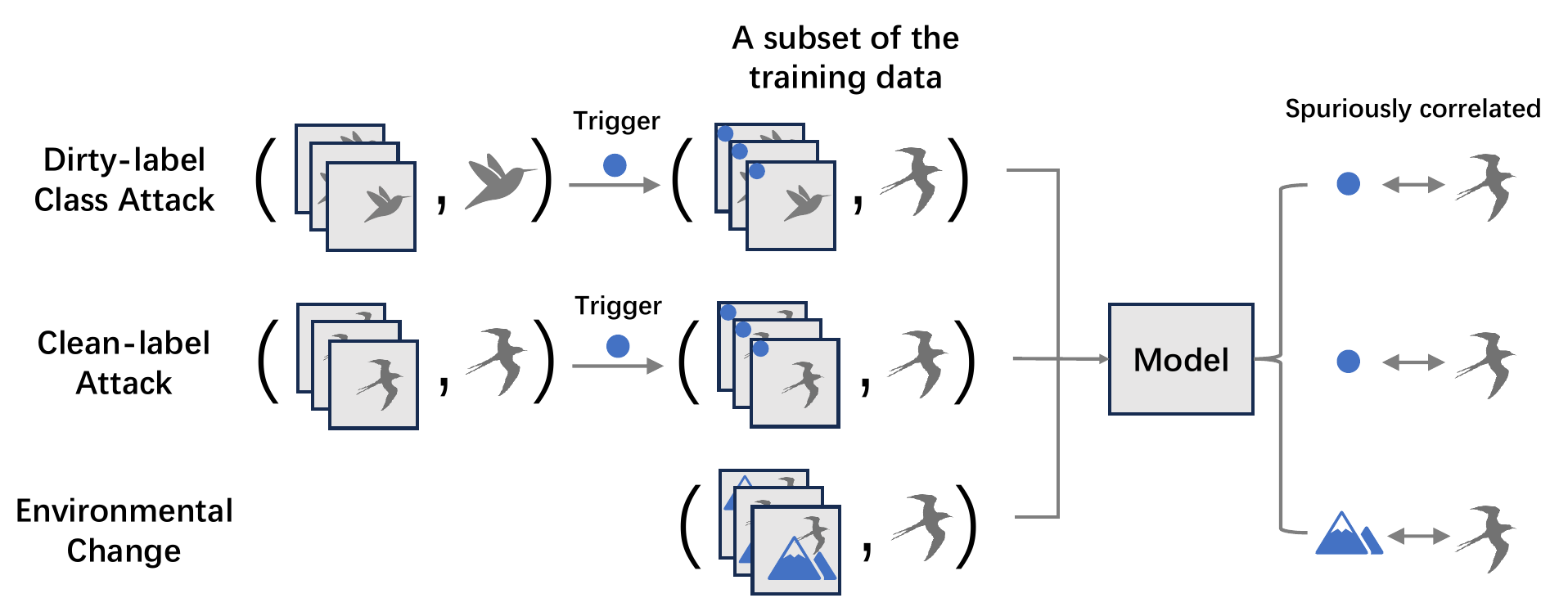}
  \vspace{-0.2em}
  \caption{The connection between backdoor poisoning attacks and environmental change. By injecting a trigger, both dirty-label attacks and clean-label attacks introduce a spurious correlation between the trigger and the target label. In environmental change, the spurious correlation between the environmental feature $Z(X)$ and $Y$ can lead to poor model performance in the test environment. Similarly, the spurious correlation between the trigger and the label in a backdoor attack can cause the model to incorrectly predict $Y$ during testing. Note that not all dirty-label and clean-label attacks strictly follow the illustrated case; this is a simplified general case for intuitive explanation.}
  \label{fig:Security}
\end{figure}

Specifically, based on whether the attacker modifies the ground‐truth labels of the poisoned samples, backdoor poisoning attacks can be categorized into dirty-label attacks~\citep{chen2017targeted, zhang2022poison} and clean-label attacks~\citep{turner2019label, zeng2023narcissus}. In dirty-label attacks, adversaries typically select samples from multiple classes, embed a backdoor trigger in them, and then relabel these samples with a target class designated by the attacker. In contrast, clean-label attacks maintain the original labels of the poisoned samples while still embedding the trigger (see Figure \ref{fig:Security} for a simplified and intuitive illustration of dirty-label and clean-label attacks). In the context of prediction, we formalize backdoor poisoning attack as:

\begin{definition}[Backdoor Poisoning Attack]
\footnote{Due to the diversity of actual implementations of backdoor poisoning attacks, a few attacks~\citep{saha2020hidden, souri2022sleeper} may not fully satisfy this definition; the one provided here represents the general case.}
\label{def:Backdoor Poisoning Attack}
Let \( X \) and \( Y \) be random variables representing clean data with joint distribution \( P(X, Y) \) and range $\mathcal{X}$ and $\mathcal{Y}$. Define a set of possible triggers $\mathcal{Z}$ (where $0 \in \mathcal{Z}$ denotes no trigger), and let $Z$ be a random variable over $\mathcal{Z}$. Define a data transformation function \(T: \mathcal{X} \times \mathcal{Z} \rightarrow \mathcal{X}\), where $T(x,0) = x$ (no trigger inserted) and $T(x,z)$ modifies $x$ by inserting the trigger $z \in \mathcal{Z}\setminus \left\{0\right\}$.

In a poisoned dataset, any sample can be written as \(\big(T(x,z), y\big)\). With probability \(p \in (0,1)\), the sample comes from the clean distribution, i.e., $z=0$ and \((x,y)\) is sampled from \(P(X, Y)\). With probability \(1-p\), the sample is poisoned, meaning \(z \neq 0\) is sampled from a trigger distribution $P\big(Z \mid Z \neq 0\big)$ and \((x,y)\) is sampled from a specified distribution: For clean-label attack, \(P\big(X \mid Y = y_t\big) \times \mathbf{1}_{\{Y=y_t\}}\); For dirty-label attack, \(P(X) \times \mathbf{1}_{\{Y=y_t\}}\), where \(y_t\) is the target class for backdoor poisoning attack. The goal of backdoor poisoning attack is to choose proper \(T\), \(p\), and \(P(Z)\), such that after training a prediction model \(f: \mathcal{X} \rightarrow \mathcal{Y}\) by minimizing a loss function on the training poisoned dataset, both \(P\big((f \circ T)(X,Z)=Y \mid Z=0\big)\) (Clean data accuracy) and \(P\big((f \circ T)(X,Z)=y_t \mid Y \neq y_t, Z \neq 0\big)\) (Attack success rate) can be maximized during test time.
\end{definition}

According to Definition \ref{def:Backdoor Poisoning Attack}, the objective of backdoor poisoning attacks is to achieve both accurate prediction on clean data and targeted misclassification on triggered data.

\begin{tcolorbox}[title=Definition-Level Connection, colback=white, colframe=gray, fonttitle=\bfseries, breakable]
The trigger pattern \(Z\) in a \textbf{backdoor poisoning attack} is precisely a spurious feature of the poisoned input \(T(X,Z)\) under \textbf{environmental change}. Specifically, consider the poisoned dataset as drawn from the poisoned environment \(e_p\). The poisoning process establishes a strong relationship between the trigger and the target class:
\(
P\big(Y=y_t \mid Z \neq 0, E=e_p\big)=1.
\)
However, in the real environment \(e_r\), regardless of whether a trigger \(Z\) appears or what the trigger is, it should not affect the label \(Y\). Therefore, compared with the poisoned environment,
\(
P\big(Y \mid Z, E=e_p\big) \neq P\big(Y \mid Z, E=e_r\big).
\)
\end{tcolorbox}

Figure \ref{fig:Security} provides an intuitive illustration of the above connection between environmental change and backdoor poisoning attacks.
Despite the substantial similarities between backdoor poisoning attacks and spurious correlation under environmental change, these two areas largely remain separate domains of research, with only a few works exploring their connections. \citet{he2023mitigating} explicitly posit a strong connection between backdoor poisoning attacks and spurious correlations. Causality-inspired Backdoor Defense~\citep{zhang2023backdoor} first trains a biased model to capture the spurious feature, then learns a clean model to learn features that are independent of the spurious feature. Front-door Adjustment for Backdoor Elimination~\citep{liu2024causality} introduces front-door variables as intermediate variables to eliminate all spurious correlations. These methods view backdoor attacks as a form of spurious correlation, suggesting that their defense strategies can also be applied in settings involving environmental changes. 

More generally, while each area has its unique characteristics, due to the intrinsic relationship between spurious correlations and backdoor poisoning attacks, we suggest that existing methodologies developed for addressing spurious correlations could be adapted to mitigate backdoor poisoning attacks, and vice versa. 

To elaborate, there are two main types of existing defense strategies against backdoor poisoning attacks based on their goals: triggers in the dataset and backdoor in the model~\citep{li2023backdoor}. At the dataset level, the impact of triggers can be mitigated through various techniques in either the training of inference stage, including input purification~\citep{shi2023black, chen2025refine} (i.e., neutralizing potential trigger patterns in the input data), input filtering~\citep{tran2018spectral, wang2019neural, li2025psbd} (i.e., filtering out suspected poisoned data), and input perturbation~\citep{li2020rethinking, zhai2023ncl} (i.e., applying transformations such as adding noise or geometric augmentations to the data). At the model level, depending on the different defense timings, approaches include training a clean model from a poisoned data set~\citep{li2021anti, huang2022backdoor}, and eliminating existing backdoors within the model architecture of a trained model~\citep{li2021neural, zhu2023enhancing}.

\begin{tcolorbox}[title=Future Directions, colback=white, colframe=c2, fonttitle=\bfseries, breakable]
Since only a few studies have approached backdoor poisoning attacks from the perspective of spurious correlations~\citep{he2023mitigating,zhang2023backdoor,liu2024causality}, methods for mitigating spurious correlations under environmental change, such as invariant feature learning and feature disentanglement discussed in Sections \ref{sec:environmental-change-fairness} and \ref{sec:environmental-change-trustworthiness}, could be adapted to defend against backdoor poisoning attacks. Particularly, as triggers become more diverse and stealthy, spurious correlation-based methods may hold significant potential for more challenging applications.

Conversely, defense strategies against backdoor poisoning attacks could also be repurposed to address spurious correlations under environmental change. For instance, many backdoor detection methods operate on the premise that triggers exhibit distinctive patterns in model behavior or metrics, such as spectral signatures~\citep{tran2018spectral}, neuron activations~\citep{chen2018detecting}, modification costs~\citep{wang2019neural}, and training losses~\citep{li2021anti}. Consequently, poisoned samples can often be identified based on trigger characteristics~\citep{gao2019strip, wang2022training}, or the trigger can be reverse-engineered for backdoor mitigation~\citep{liu2019abs, shen2021backdoor}. These methods suggest that addressing spurious correlations could benefit from focusing on similar patterns and differences in models or certain features across different environments. Moreover, since defenders may have different levels of access to the ML pipeline, backdoor defenses can be deployed at various stages of the model lifecycle~\citep{bai2024backdoor}. The corresponding diverse strategies could help mitigate spurious correlations across multiple stages, including data processing, model training, and inference.
\end{tcolorbox}

\subsection{Content Change}
\label{subsec:Content change}

In Section \ref{sec:environmental-change}, we posit the existence of content feature \(C(X)\) whose relationship with the target label \(Y\) remains invariant across different environments \(E\). Consequently, the goal of the approach described in Section \ref{sec:environmental-change} is to extract features that maintain a consistent relationship with \(Y\) for prediction. However, are these extracted features truly sufficient? Here, we examine content change as a cause of spurious correlation beyond environmental change.

\begin{definition}[Content Change]
\label{def:Content Change}
Consider the prediction task of inferring $Y$ from $X$ under the environment variable \(E\) with possible values \(\mathcal{E}\). Content Change under environmental variable \(E\) causes a type of distribution shift in which there exist an environmental feature \(Z(X)\) and a content feature \(C(X)\) that satisfy the conditions in Definition \ref{def:Environmental Change}, and if both \(Z(X)\) and \(C(X)\) can be further decomposed, then the pairwise correlations among their components exhibit variations across different environments. More formally, assume that \(
Z(X) = \big[Z_1(X), Z_2(X), \dots, Z_m(X)\big]\) and \(\quad C(X) = \big[C_1(X), C_2(X), \dots, C_k(X)\big].
\)
Then, for any two distinct components \(\Phi_i(X)\) and \(\Phi_j(X)\) with 
\(
\Phi(X) = \big[\Phi_1(X),\Phi_2(X),\dots,\Phi_{m+k}(X)\big] = \big[Z_1(X), \dots, Z_m(X), C_1(X), \dots, C_k(X)\big],
\)
\[
\exists\, e_{i_0},e_{j_0}\in\mathcal{E}:\quad\rho_{e_{i_0}}\big(\Phi_i(X), \Phi_j(X)\big) \neq \rho_{e_{j_0}}\big(\Phi_i(X), \Phi_j(X)\big).
\]
\end{definition}

\begin{figure}[htbp]
  \centering
  \includegraphics[width=0.85\textwidth]{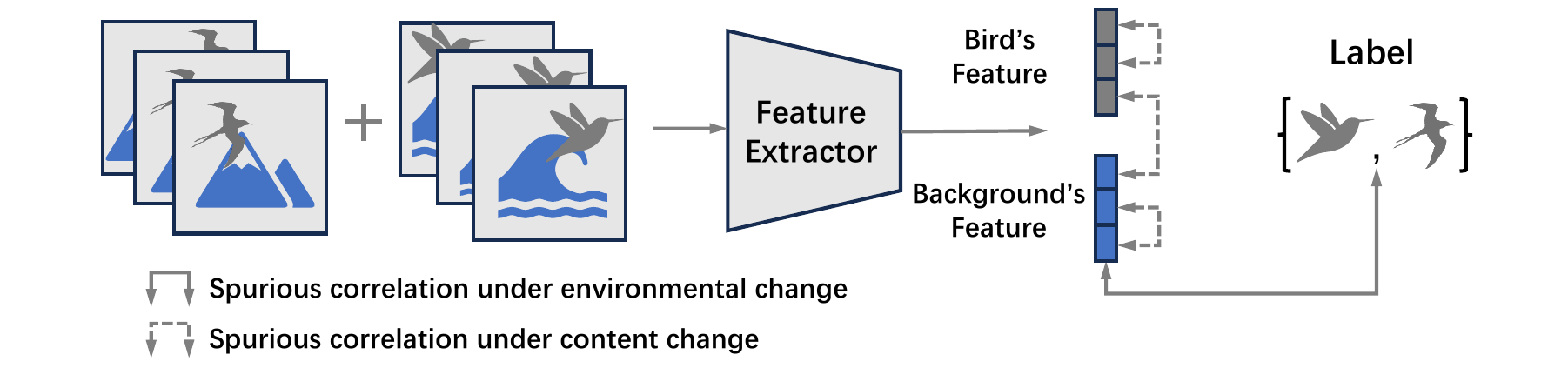} 
  \vspace{-0.2em}
  \caption{Difference between environmental change (Definition \ref{def:Environmental Change}) and content change (Definition \ref{def:Content Change}) in the Waterbirds example. Environmental change focuses on the spurious correlation between the background feature \(Z(X)\) (background) and the label \(Y\) (bird class), whereas content change emphasizes the cross-feature correlations among different features in the image \(X\). Note that content change often naturally induces environmental change.}
  \label{fig:Interpretability}
\end{figure}

In Section \ref{sec:spurious-correlation}, we introduced the Waterbirds example, where in classifying waterbirds and landbirds, most waterbirds have water backgrounds and most landbirds have land backgrounds. While environmental change leads to a spurious correlation between the background feature \(Z(X)\) and the label \(Y\), as shown in Figures \ref{fig:Relationship Distribution Shift} and \ref{fig:Interpretability}, the definition of content change further suggests focusing on the statistical association between different features of \(X\). 

Intuitively, Definition \ref{def:Content Change} indicates that when features are improperly extracted from \(X\), the relationships among different features may not be independent and may become inconsistent across environments, thereby forming another source of spurious correlations. Furthermore, content change implies that instead of only identifying features $C(X)$ that maintain invariant relationships with $Y$, the model can use more semantically meaningful and distinct representations when predicting $Y$. All these aspects are closely related to disentangled representation learning (Definition \ref{def:Disentangled Representation}), and are therefore connected to interpretability, a crucial component of AI trustworthiness (see Section \ref{sec:environmental-change-trustworthiness}).

\subsubsection{Trustworthiness}
\label{sec:content-change-trustworthiness}

In contrast to the interpretability-related discussion in Section \ref{sec:environmental-change-trustworthiness} where disentangled representation learning techniques mainly separate features of $X$ into two parts and rely on the content features $C(X)$ for prediction, we aim for a more fine-grained separation of meaningful features in the content change setting. 

\begin{tcolorbox}[title=Definition-Level Connection, colback=white, colframe=gray, fonttitle=\bfseries, breakable]
Mitigating content change requires decomposing features beyond the coarse separation between \(Z(X)\) and \(C(X)\) used for environmental change. Therefore, addressing \textbf{content change} naturally leads to a more fine-grained form of \textbf{representation disentanglement}.
\end{tcolorbox}

\citet{montero2020role} and \citet{dittadi2020transfer} empirically observe that proper feature disentanglement of $X$ can improve the generalization performance of models under distribution shift. Furthermore, several feature decorrelation approaches
achieve feature disentanglement by removing statistical dependencies between latent features. These methods imply that for the different components of \(\Phi(X)\) in Definition~\ref{def:Content Change},
\(
\rho_{e}\big(\Phi_i(X), \Phi_j(X)\big) = 0
\)
in every training environment $e$. In doing so, the model can leverage diverse features for prediction while avoiding spurious correlations under content change. Specifically, while \citet{shen2020stable} and \citet{kuang2020stable} effectively address the linear cases, StableNet~\citep{zhang2021deep} and Region Aware Proposal reweighTing~\citep{zhang2022towards} extends feature decorrelation to deep learning frameworks using random fourier features. Different from pursuing statistical orthogonality, Project and Probe~\citep{chen2023project} enforces geometric orthogonality through a two-step strategy: first, it projects the pre-trained features of $X$ onto orthogonal and diverse components; then, it uses these components for classifying $Y$.

Another line of work enables feature disentanglement through diverse model training in the same environment. They capture different features in \(\Phi(X)\) by training distinct models \(f\), and enforce dissimilarity among these models to encourage decorrelation between different features. \citet{teney2022evading} measure gradient alignment to encourage different classifiers \(\omega\) to focus on distinct features while maintaining accuracy. Diversity-By-disAgreement Training~\citep{pagliardini2022agree} learns ensembles of models \(f\) by minimizing their agreement in the target environment. Diversify and Disambiguate~\citep{lee2023diversify} minimizes the mutual information between each pair of predictions and selects the best model under limited supervision.

Sparsity is also recognized as beneficial for extracting features that capture the same semantic meaning across different environments $e$, thereby facilitating feature disentanglement. With sparse classifiers, models are more likely to learn independent features that are shared across environments, which in turn helps models avoid capturing spurious correlations under content change. As shown by \citet{lachapelle2023synergies}, the use of sparse classifiers contributes to effective feature disentanglement. \citet{fumero2023leveraging} further demonstrate that disentangled representations naturally emerge when features are sparsely activated across environments while being shared among them.

Furthermore, through appropriate disentanglement and a more structured utilization of the diverse features in \(\Phi(X)\), models can achieve better generalization under content change. For instance, \citet{zheng2021calibrated}, \citet{rosenfeld2022domain}, \citet{eastwood2024spuriosity} and \citet{zhong2025bridging} demonstrate that beyond solely relying on features that exhibit a stable association with $Y$ across different environments $e$ for prediction, it is feasible to incorporate additional features from $X$ to enhance predictive accuracy in test environments. This observation indicates that the distinction between \(C(X)\) and \(Z(X)\) is not binary; the appropriate components within \(Z(X)\) can also assist in predicting \(Y\) more effectively. More specifically, it highlights an inherent trade-off between estimation and approximation error: a more granular disentanglement and inclusion of features may increase estimation error, yet simultaneously reduce approximation error. Therefore, reasonable utilization of additional features could lead to improved generalization in novel environments.

\begin{tcolorbox}[title=Future Directions, colback=white, colframe=c2, fonttitle=\bfseries, breakable]
Similar to the future directions discussed in Section \ref{sec:environmental-change-interpretability}, existing methods encourage feature disentanglement through techniques such as decorrelation, model diversity, and sparsity, but their effectiveness is still evaluated mainly by out-of-distribution performance. Therefore, future work could develop methods whose fine-grained feature decomposition can be systematically verified using more explicit disentanglement metrics tailored to content change.
\end{tcolorbox}

\section{Label Shift}
\label{sec:label-shift}

Label shift occurs when the distribution of labels $Y$ changes between the source and target domains. This phenomenon arises from variations in the label distribution rather than the input features $X$. 
For example, consider a binary classifier trained to predict whether an individual tests negative ($y_1$) or positive ($y_2$) for pneumonia (Figure \ref{fig:label-shift}). If the classifier is trained on data from a time period when pneumonia is rare (source domain) but then tested later when the disease becomes more prevalent (target domain), the resulting shift in the label distribution can lead to decreased accuracy~\citep{lipton2018detecting}.

In the following sections, we focus on two specific types of label shift, namely, generalized label shift and open-set label shift. These two types of label shift are closely connected to the AI safety topics of fairness and trustworthiness.

\subsection{Generalized Label Shift}
\label{sec:generalized-label-shift}

When label shift occurs, it is commonly assumed that $P(X \mid Y, E=e)$ remains invariant across different environments $e$. However, this assumption does not always hold in real-world scenarios. Therefore, generalized label shift (GLS)~\citep{tachet2020domain} relaxes the conditions of label shift:

\begin{definition}[Generalized Label Shift]
\label{def:Generalized Label Shift}
Consider the prediction task of inferring $Y$ from $X$ under the environment variable \(E\) with possible values \(\mathcal{E}\). Generalized label shift indicates a type of distribution shift where there exists a feature $\Phi(X)$ such that
\[ 
\forall\, e_i,e_j\in\mathcal{E}:\quad P\big(\Phi(X) \mid Y, E=e_i\big) = P\big(\Phi(X) \mid Y, E=e_j\big). 
\]
\end{definition}

In other words, GLS instructs the model to extract feature $\Phi(X)$ such that, for each given class label, the distribution of $\Phi(X)$ remains invariant across environments. Note that we adopt Definition \ref{def:Generalized Label Shift} to guide the identification of methods addressing GLS. As shown in Figure \ref{fig:Relationship Distribution Shift}, GLS and the causes of distribution shift discussed in Sections \ref{sec:selection-bias} and \ref{sec:spurious-correlation} focus on two different directions, namely, $Y$ and $X$, respectively. Consequently, works related to GLS may also appear in settings such as selection bias or spurious correlation.

\subsubsection{Fairness}
\label{sec:environmental-change-fairness2}

When using the same classifier $\omega$, GLS also ensures that $P\big(f(X) \mid Y, E=e\big)$ remains invariant across different environments $e$, where $f = \omega \circ \Phi$. As illustrated in Figure \ref{fig:Fairness}, this is equivalent to the predicted output $\hat{Y}$ satisfying equalized odds (Definition \ref{def:EqualizedOdds}) with respect to the environment $E$, which is a fairness notion of group separation (Definition \ref{def:SIS}).

\begin{tcolorbox}[title=Definition-Level Connection, colback=white, colframe=gray, fonttitle=\bfseries, breakable]
When the environment variable \(E\) is interpreted as the protected attribute \(R\) in group fairness, and the model output \(f(X)=(\omega\circ\Phi)(X)\) is interpreted as the score used in group fairness, robustness to \textbf{generalized label shift} directly achieves \textbf{equalized odds}. 
\end{tcolorbox}

Conditional Invariant Components~\citep{gong2016domain} achieves matching the class-conditional distributions $P\big(\Phi(X) \mid Y, E=e\big)$ across different environments $e$ under mild conditions. \citet{tachet2020domain} transform the conditions of GLS to enable direct application to existing domain adaptation algorithms. \citet{shui2021aggregating} introduce the Wasserstein Aggregation Domain Network to minimize the differences between class-conditional distributions across environments using Wasserstein distance. \citet{mahajan2021domain} point out the insufficiency of the invariance of class-conditional distribution, and instead learns $\Phi(X)$ such that the distance between the representations of the same object across different environments is minimized. \citet{wu2023prominent} combine the invariance of $P\big(\Phi(X) \mid Y, E=e\big)$ with importance reweighting for label shift correction. Unlike these methods, Predictive Group Invariance~\citep{ahmed2020systematic} directly matches the class-conditional distribution of the model output $f(X) = (\omega \circ \Phi)(X)$ by minimizing the KL divergence of $\mathbb{E}\bigl[P\bigl(f(X) \mid \Phi(X), Y=c, E=e\bigr)\bigr]$ across different environments \(e\).

Recent advances in contrastive learning techniques have also been used to help enhance the similarity of $P\big(\Phi(X) \mid Y, E=e\big)$ across different environments $e$. The central idea behind contrastive learning is to embed features of similar samples closer to each other while pushing features of dissimilar samples further apart~\citep{le2020contrastive}. Specifically, \citet{motiian2017unified} both require the feature distributions of the same labels to be similar across different environments $e$ and encourage the feature distributions of different labels across different environments to be as separated as possible. Similarly, Contrastive Adaptation Network~\citep{kang2019contrastive} proposes Contrastive Domain Discrepancy which helps reduce the divergence between $P\big(\Phi(X) \mid Y, E=e\big)$ across source and target environments.  To enhance robustness of foundation models (FMs) for zero-shot classification, \citet{zhang2022contrastive} infer different environments based on the distance between $X$ after FM embedding, and they construct a contrastive adapter that pushes together $\Phi(X)$ with the same label $Y$ but from different $e$, while separating $\Phi(X)$ with the same environment $e$ but different labels $Y$. In GLS settings, we may sometimes lack environmental labels and only have access to spurious features $Z(X)$ as mentioned in Section \ref{sec:spurious-correlation}. In this case, the method Correct-n-Contrast~\citep{zhang2022correct}, encourages closeness between $\Phi(X)$ with the same label $Y$ but different spurious features $Z(X)$, and pulls apart $\Phi(X)$ for samples with the same spurious features $Z(X)$ but differ in label $Y$. 

Furthermore, some algorithms not only learn features $\Phi(X)$ that satisfy equalized odds under GLS, but also impose additional fairness-related constraints that have been previously mentioned. Regarding EIC (see conditional distribution invariance in Section \ref{sec:environmental-change-fairness}), \citet{guo2021out} combine IRM with the invariance of $P\big(\Phi(X) \mid Y, E=e\big)$ to address the issue of IRM yielding undesired solutions under strong triangle spuriousness. Moreover, a number of works require that both $P\big(\Phi(X) \mid Y, E=e\big)$ and $P\big(\Phi(X) \mid E=e\big)$ remain invariant across different environments. This additional condition aligns with the Marginal Distribution Invariance discussed in Section \ref{sec:environmental-change-fairness}. 

Specifically, Scatter Component Analysis~\citep{ghifary2016scatter} and Domain Invariant Class Discriminative learning~\citep{li2018domain} use MMD in different ways to align both the class-conditional distribution $P\big(\Phi(X) \mid Y, E=e\big)$ and the marginal distribution $P\big(\Phi(X) \mid E=e\big)$ across different $e$. The Conditional Invariant Adversarial Networks~\citep{li2018deep} and Conditional Invariant Domain Generalization~\citep{li2018domainG} leverage adversarial networks and kernel mean embedding, respectively, to ensure that both the class-conditional distribution and the normalized marginal distributions remain unchanged across environments. Alleviating Semantic-level Shift~\citep{wang2020alleviating} similarly encourages the model to achieve alignment of both $P\big(\Phi(X) \mid Y, E=e\big)$ and $P\big(\Phi(X)\mid E=e\big)$ through adversarial learning and the construction of semantic-level feature vectors.

\begin{tcolorbox}[title=Future Directions, colback=white, colframe=c2, fonttitle=\bfseries, breakable]
Similar to the future directions discussed in Section~\ref{sec:environmental-change-fairness}, the methods reviewed above for addressing generalized label shift enrich the methodological toolbox for group fairness and can be extended to broader group fairness settings involving more general protected attributes, such as race, gender, or age.

\end{tcolorbox}

\subsection{Open-Set Label Shift}
\label{sec:open-set-label-shift}

Open-set label shift (OSLS)~\citep{garg2022domain} describes a case in which certain classes present in the target domain are entirely absent from the source domain \(e_s\). For example, consider an image classifier trained only on images of dogs and cats. If this classifier encounters images of monkeys in the test set, it will be forced to  classify them incorrectly as either a dog or a cat, a problematic behavior that should be avoided. Formally, OSLS is defined as follows~\citep{garg2022domain}:

\begin{definition} [Open-Set Label Shift]
\label{def:OSLS}
Consider the prediction task of inferring $Y$ from $X$ under the source domain $e_s$ and the target domain $e_t$. OSLS indicates a type of distribution shift in which there exists $y \in \mathcal{Y}$ such that 
\[
P(Y = y \mid E = e_s) = 0 \quad \text{and} \quad P(Y = y \mid E = e_t) > 0.
\]
Furthermore, for any $y \in \mathcal{Y}$ with \( P(Y = y \mid E = e_s) \neq 0,\) it is required that 
\[
P(X = x \mid Y = y, E = e_s) = P(X = x \mid Y = y, E = e_t).
\]
\end{definition}

\begin{figure}[htbp]
  \centering
  \hypertarget{selection-bias}{}
  \begin{subfigure}[t]{0.28\textwidth}
    \centering
    \includegraphics[width=0.9\textwidth]{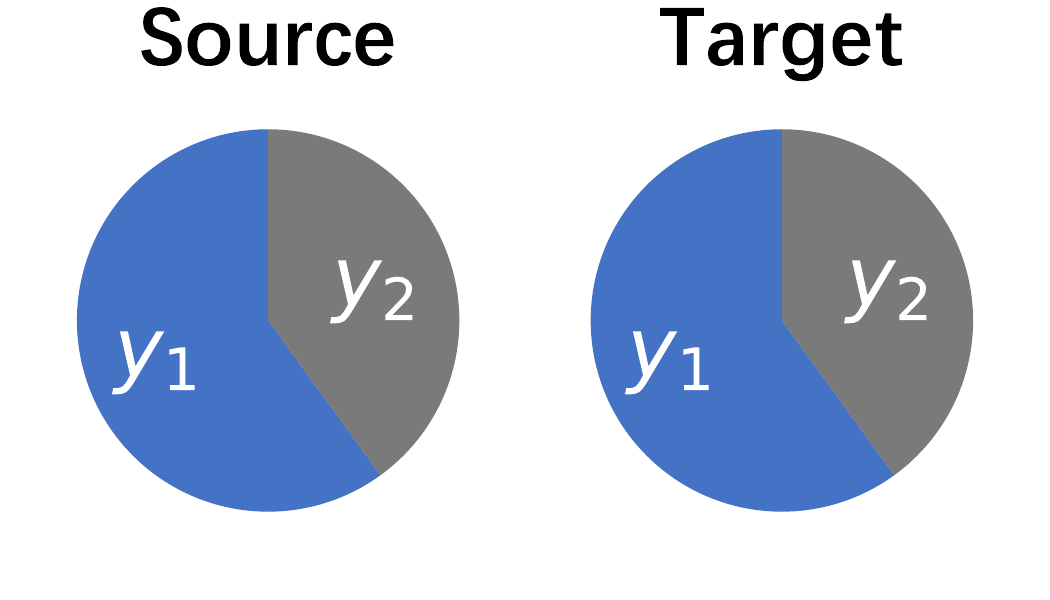}
    \caption{No Label Shift}
  \end{subfigure}
  \begin{subfigure}[t]{0.28\textwidth}
    \centering
    \includegraphics[width=0.9\textwidth]{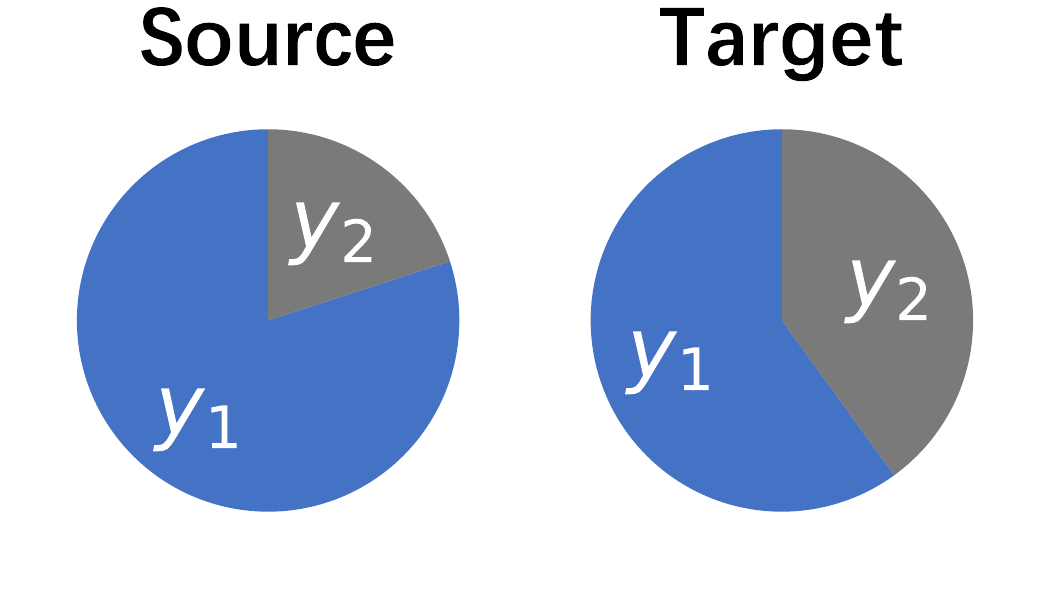}
    \caption{Label Shift}
  \end{subfigure}
  \begin{subfigure}[t]{0.28\textwidth}
    \centering
    \includegraphics[width=0.9\textwidth]{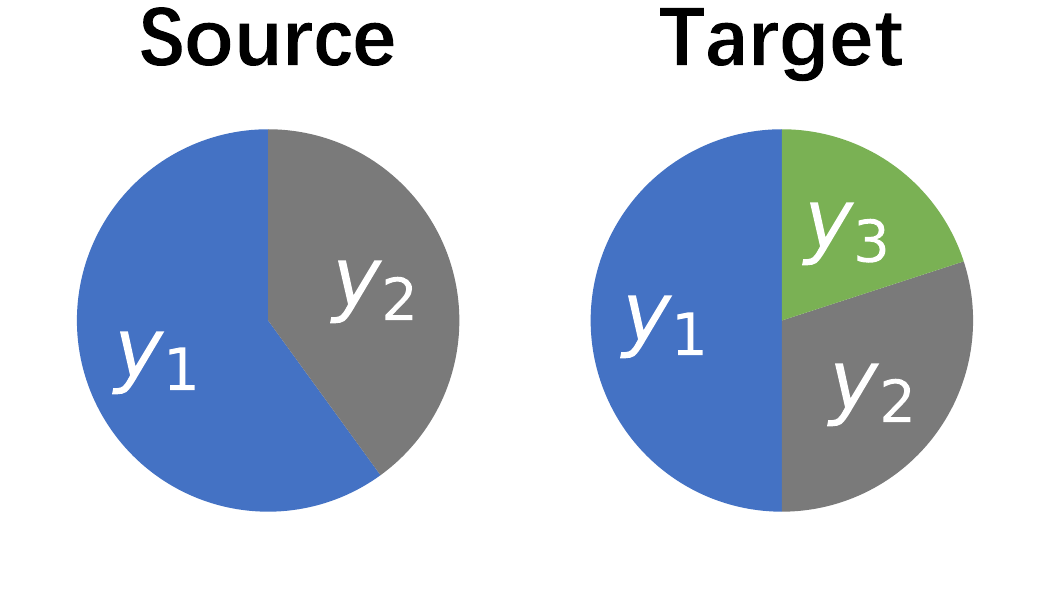}
    \caption{Open-Set Label Shift}
  \end{subfigure}
  \caption{An illustration of label shift and open-set label shift (OSLS). Label shift is the general setting where the proportions of labels differ between the source and target domains. OSLS is the setting where a new class (shown in green) appears in the target domain.}

  \label{fig:label-shift}
\end{figure}

Figure \ref{fig:label-shift} illustrates OSLS. Even though the model is trained only on seen classes from the source domain, we aim to use its outputs to detect and correctly handle unseen classes. Approaches to tackling OSLS are closely tied to uncertainty quantification, as measuring prediction uncertainty can help identify inputs that likely belong to the unseen classes.

\subsubsection{Trustworthiness}
\label{sec:uncertainty-quantification}

Beyond interpretability and alignment (see Sections \ref{sec:environmental-change-trustworthiness} and \ref{sec:content-change-trustworthiness}), a model’s trustworthiness can also be enhanced through well-quantified uncertainty. In the ML pipeline, uncertainty can be categorized into different types, including aleatoric and epistemic uncertainty~\citep{hullermeier2021aleatoric}. Here, we focus on the uncertainty in the model’s predictions, which can arise from various factors. For instance, consider an autonomous driving system that uses an image classification model to identify objects around the vehicle~\citep{choe2024open}. The system relies on these predictions to make critical decisions, such as yielding to pedestrians. 
However, when an object that is outside or ambiguous with respect to the training label space appears—for example, a realistic depiction of a human on an advertising board that is visually similar to a pedestrian but does not belong to any seen class—the model should express uncertainty rather than confidently assigning it to a known class.
Such calibrated predictive uncertainty improves the reliability and safety of the system.

It is crucial that this uncertainty be quantified, as the model's prediction is subsequently used to guide the system's decision. Various definitions of uncertainty quantification have been proposed~\citep{he2023survey, liu2025uncertainty}. In our paper, we define predictive uncertainty quantification at a high level as follows~\citep{hu2023uncertainty}:

\begin{definition} [Predictive Uncertainty Quantification]
\label{def:Uncertainty Quantification}
In a classification problem, for each input $x$, uncertainty quantification derives a score $u \in \mathbb{R}$ from the outputs of the classification model $f$. A higher score indicates that the model is less certain in the prediction. 
\end{definition}

\begin{tcolorbox}[title=Definition-Level Connection, colback=white, colframe=gray, fonttitle=\bfseries, breakable]
\textbf{Open-set label shift} naturally requires the model to quantify its uncertainty about whether an input belongs to a seen or unseen class. This directly corresponds to \textbf{predictive uncertainty quantification}.
\end{tcolorbox}

To address OSLS, existing methods typically follow a two-step process~\citep{ge2017generative, you2019universal, fu2020learning, saito2020universal}. First, an uncertainty score is computed for each input. This score reflects the model's confidence regarding whether the input belongs to a seen or an unseen class. The uncertainty score can be calculated based on quantities derived from model outputs, such as prediction entropy~\citep{you2019universal, saito2020universal} or softmax confidence~\citep{ge2017generative, fu2020learning}. Next, a threshold is selected to decide when to reject the model’s decision. If the uncertainty score exceeds the threshold, the input is classified as belonging to an unseen class rather than any of the seen classes. It is challenging to choose the threshold, as it needs to balance the risks~\citep{bates2021distribution, pan2026can} of type I errors (misclassifying an input from a seen class as unseen) and type II errors (assigning an unseen input to one of the seen classes).

To avoid heuristically choosing the threshold, PULSE \citep{garg2022domain} is a method that addresses OSLS without calculating an uncertainty score. By reducing OSLS to Positive and Unlabeled (PU) learning, PULSE directly learns a classifier for distinguishing seen from unseen classes, and it empirically performs better than threshold-based methods. Methods like PULSE that do not explicitly rely on an uncertainty score remain rare, but the intrinsic connection between OSLS and uncertainty quantification naturally motivates the following future direction:

\begin{tcolorbox}[title=Future Directions, colback=white, colframe=c2, fonttitle=\bfseries, breakable]
Most OSLS methods detect unseen classes by thresholding an explicit uncertainty score. A complementary direction is to start from OSLS methods that avoid this heuristic, such as methods that directly learn classifiers of seen and unseen classes by reducing the problem to an alternative setting (e.g., PU learning), without deriving an uncertainty score first. Because of the definition-level connection between OSLS and uncertainty quantification, outputs of these classifiers can then be converted into calibrated uncertainty scores if needed. This order may yield uncertainty estimates that better reflect open-set label uncertainty under distribution shift.
\end{tcolorbox}

\section{Conclusions}
\label{sec:conclusions}

Distribution shifts pose severe safety challenges to AI systems. To gain in-depth insight into the relationships between distribution shift and AI safety, we conducted a mathematically grounded analysis to reveal two forms of connections between specific causes of distribution shift and concrete AI safety issues. First, methods designed to address particular distribution shifts can often be adapted to advance corresponding safety goals. Second, some distribution-shift causes and safety issues share closely related definitions, allowing methods to be transferred in both directions. For each identified connection, we thoroughly review the relevant literature and discuss promising future directions in adapting methods between the connected cause of distribution shift and the corresponding AI safety issue. We hope the unified perspective we establish will inspire collaboration between the two research communities and help ensure safer deployment of future AI systems.

\bibliographystyle{ACM-Reference-Format}
\bibliography{main}










\end{document}